\documentclass{article}

\usepackage{microtype}
\usepackage{graphicx}
\usepackage{subfigure}
\usepackage{booktabs} % for professional tables

\usepackage{hyperref}

\usepackage[accepted]{icml2024}

\usepackage{amsmath}
\usepackage{amssymb}
\usepackage{mathtools}
\usepackage{amsthm}

\usepackage[capitalize,noabbrev]{cleveref}

\theoremstyle{plain}

\theoremstyle{definition}

\theoremstyle{remark}

\usepackage[textsize=tiny]{todonotes}

\usepackage[T1]{fontenc}
\DeclareMathOperator*{\argmax}{arg\,max}
\usepackage{subcaption}
\usepackage{comment}
\usepackage{bm}
\usepackage{balance}
\usepackage{multirow}

\icmltitlerunning{Individual Contributions as Intrinsic Exploration Scaffolds for Multi-agent Reinforcement Learning}

\begin{document}

\twocolumn[
\icmltitle{Individual Contributions as Intrinsic Exploration Scaffolds for Multi-agent Reinforcement Learning}



\icmlsetsymbol{equal}{*}

\begin{icmlauthorlist}
\icmlauthor{Xinran Li}{hkust}
\icmlauthor{Zifan Liu}{hkust}
\icmlauthor{Shibo Chen}{hkust}
\icmlauthor{Jun Zhang}{hkust}
% \icmlauthor{Firstname3 Lastname3}{comp}
\end{icmlauthorlist}

\icmlaffiliation{hkust}{Department of Electronic and Computer Engineering, The Hong Kong University of Science and Technology, Hong Kong SAR, China}

\icmlcorrespondingauthor{Shibo Chen}{eeshibochen@ust.hk}

\icmlkeywords{Multi-agent Reinforcement Learning, Exploration}

\vskip 0.3in
]



\printAffiliationsAndNotice{}  % leave blank if no need to mention equal contribution

\begin{abstract}
In multi-agent reinforcement learning (MARL), effective exploration is critical, especially in sparse reward environments.
Although introducing global intrinsic rewards can foster exploration in such settings, it often complicates credit assignment among agents. To address this difficulty, we propose Individual Contributions as intrinsic Exploration Scaffolds (ICES), a novel approach to motivate exploration by assessing each agent's contribution from a global view. In particular, ICES constructs \emph{exploration scaffolds} with Bayesian surprise, leveraging global transition information during centralized training. These scaffolds, used only in training, help to guide individual agents towards actions that significantly impact the global latent state transitions. Additionally, ICES separates exploration policies from exploitation policies, enabling the former to utilize privileged global information during training. Extensive experiments on cooperative benchmark tasks with sparse rewards, including Google Research Football (GRF) and StarCraft Multi-agent Challenge (SMAC), demonstrate that ICES exhibits superior exploration capabilities compared with baselines. The code is publicly available at \url{https://github.com/LXXXXR/ICES}.
\end{abstract}

\section{Introduction}

Multi-agent reinforcement learning (MARL) has recently gained significant interest in the research community, primarily due to its applicability across a diverse range of practical scenarios. Numerous real-world applications are multi-agent in nature, ranging from resource allocation~\citep{marl_dis} and package logistics~\citep{marl_delivery} to emergency response operations~\citep{marl_rescue} and robotic control systems~\citep{marl_robot_swamy2020scaled}. The multi-agent settings introduce unique challenges beyond the single agent reinforcement learning (RL), such as the need to address non-stationarity and partial observability~\citep{marl_survey}, as well as the complexities involved in credit assignment~\citep{COMA}. 

Despite the progress made by state-of-the-art algorithms like MADDPG~\citep{MADDPG}, QMIX~\citep{QMIX} and MAPPO~\citep{MAPPO}, which leverage the centralized training decentralized execution (CTDE) paradigm, a significant limitation arises in environments with sparse rewards. Sparse rewards, common in real-world applications, present a substantial challenge for policy exploration as they provide limited guidance during training. Classical exploration methods, such as $\epsilon$-greedy, struggle in these environments, primarily due to the exponentially growing state space and the necessity for coordinated exploration among agents~\citep{CMAE}. To address sparse rewards in MARL, recent approaches have focused on augmenting extrinsic rewards with global intrinsic rewards. These intrinsic rewards are typically designed to foster cooperation~\citep{EITI} or diversity~\citep{CDS}. While showing promise, these methods suffer from one obstacle: the non-stationary nature of intrinsic rewards during training~\citep{curiosity_analysis} introduces additional complications in credit assignment. Furthermore, balancing intrinsic and extrinsic rewards often demands considerable tunning effort, particularly in the absence of prior knowledge about the extrinsic reward functions~\citep{AIRS}. 

To address the aforementioned challenges and improve performance in MARL with sparse rewards, we propose a new exploration method, named Individual Contribution as Intrinsic Exploration Scaffolds (ICES). The key idea is to take advantage of the CTDE paradigm and utilize global information available during the training to construct intrinsic scaffolds that guide multi-agent exploration. These intrinsic scaffolds are specifically designed to encourage individual actions that have a significant influence on the underlying global latent state transitions, thus promoting cooperative exploration without the need to learn intrinsic credit assignment. Furthermore, these scaffolds, akin to physical scaffolds in construction, will be dismantled after training to prevent any effect on execution latency.

In particular, we make two key technical contributions. Firstly, we shift the focus from global intrinsic rewards to individual contributions as the primary motivation for agent exploration. This approach effectively circumvents the complexities involved in credit assignment for global intrinsic rewards. To encourage agents to perform cooperative exploration, we capitalize on the centralized training to estimate the Bayesian surprise related to agents' actions, quantifying their individual contributions. This is achieved by employing a conditional variational autoencoder (CVAE) with two encoders. Secondly, we optimize exploration and exploitation policies separately with distinct RL algorithms. In this way, the exploration policies can be granted access to privileged information, such as global observations, which helps to alleviate the non-stationarity challenge. Importantly, these exploration policies serve as temporary scaffolds, and do not intrude on the decentralized nature of the execution phase.

We evaluate the proposed ICES on two benchmark environments: Google Research Football (GRF) and StarCraft Multi-agent Challenge (SMAC), under sparse reward settings. The empirical results and comprehensive ablation studies demonstrate ICES's superior exploration capabilities, notably in convergence speed and final win rates, when compared with existing baselines.

\section{Background}
In this section, we briefly introduce the fully cooperative multi-agent task considered in this work and provide essential background on CVAEs, which will be utilized for constructing meaningful individual contribution assessment. Then, we provide an overview of related works on multi-agent exploration.

\subsection{Problem Setting}
\textbf{Decentralized Partially Observable Markov Decision Process (Dec-POMDP): }
We consider a fully cooperative partially observable multi-agent task modeled as a decentralized partially observable Markov decision process (Dec-POMDP)~\citep{pomdp_oliehoek2016concise}. The Dec-POMDP is defined by a tuple $\mathcal{M} = \langle \mathcal{S}, U, P, R, \Omega, O, n, \gamma \rangle$, where $n$ denotes the number of agents and $\gamma \in (0, 1]$ is the discount factor that balances the trade-off between immediate and long-term rewards. 

At timestep $t$, with the global observation $s \in \mathcal{S}$, agent $i$ receives a local observation $ o_i \in \Omega$ drawn from the observation function $O(s, i)$. Subsequently, the agent selects an action $u_i \in U$ based on its local policy $\pi_i$. These individual actions collectively form a joint action $\boldsymbol{u} \in U^n$, leading to a transition to the next global observation $s' \sim P(s'| s, \boldsymbol{u})$ and yielding a global reward $r = R(s, \boldsymbol{u})$. For clarity, we refer to this global reward as the extrinsic reward $r_\text{ext}$, distinguishing it from the agents' intrinsic motivations. Each agent keeps a local action-observation history denoted as $h_i  \in (\Omega \times U)$. The team objective is to learn the policies that maximize the expected discounted accumulated reward $G_t = \sum_t \gamma^t r^t$.

\subsection{Conditional Variational Autoencoders (CVAEs)}
CVAEs~\citep{CVAE} extend variational autoencoders (VAEs) to model conditional distributions, adept at handling scenarios where the mapping from input to output is not one-to-one, but rather one-to-many~\citep{CVAE}. The generation process in a CVAE is as follows: given an observation $\bold{x}$, a latent variable $\bold{z}$ is sampled from the prior distribution $p_\theta(\bold{z}|\bold{x})$, and the output $\bold{y}$ is generated from the conditional distribution $p_\theta(\bold{y}|\bold{x}, \bold{z})$. The objective is to maximize the conditional log-likelihood, which is intractable in practice. Therefore, the variational lower bound is maximized instead, which is expressed as:
\begin{align}
    \mathcal{L}_\text{CVAE} (\bold{x}, \bold{y}; \theta, \phi) = D_{\text{KL}} &\left[ q_\phi(\bold{z}|\bold{x}, \bold{y}) \parallel p_\theta(\bold{z}|\bold{x})  \right] \nonumber\\
    &+ \mathbb{E}_{q_\phi(\bold{z}|\bold{x}, \bold{y})}\left[ \log p(\bold{y}|\bold{x}, \bold{z})\right],
\end{align}
where $D_{\text{KL}}$ denotes the Kullback–Leibler (KL) divergence and $q_\phi(\bold{z}|\bold{x}, \bold{y})$ is an approximation of the true posterior.

\subsection{Related Works}
Besides adapting exploration techniques from single-agent RL, considerable research has focused on developing exploration methods tailored to multi-agent settings. We categorize these efforts into two broad types: global-level and agent-level exploration.

\textbf{Global-level Exploration:}
Research in this domain aims to encourage exploration in the global space. For instance, MAVEN~\cite{MAVEN} integrates hierarchical control by introducing a latent space to guide exploration. Studies by \citet{CMAE}, \citet{subspace_expl} and \citet{fox} focus on reducing the exploration space by identifying key subspaces. MAGIC~\citep{MAGIC} adopts goal-oriented exploration for multi-stage tasks. Other approaches emphasize encouraging desirable collective behaviors, such as the work by \citet{synergistic_behavior} that emphasizes fostering synergistic behaviors among agents, and LAIES~\citep{LAIES}, which avoids lazy agents by encouraging diligence. Additionally, methods like EMC~\citep{EMC} and MASER~\citep{MASER} seek to enhance sample efficiency by effectively utilizing existing experiences in replay buffers,  either by replaying high-reward sequences or creating subgoals for cooperative exploration.

\textbf{Agent-level Exploration:}
This category of research incorporates specific objectives at the individual agent level. EITI and EDTI~\citep{EITI} focus on maximizing the mutual influence among agents' state transitions and values. \citet{CDS} promotes diverse behaviors among agents and \citet{policy_diversity} proposes to encourage diverse joint policy compared to historical ones. SMMAE~\citep{SMMAE} fosters individual curiosity, while ADER~\citep{ADER} introduces an adaptive entropy-regularization scheme to allow varied levels of exploration across agents.

While global-level exploration aids in fostering cooperative behaviors, the integration of global extrinsic and intrinsic rewards often complicates credit assignment, potentially hindering algorithm performance. Some methods~\citep{MAGIC,LAIES} also rely on parsing the global observation, and thus require specific domain knowledge, which thereby limits their applicability. In contrast, agent-level exploration offers a more straightforward approach but may result in less coordinated actions among agents. Our method seeks to combine the advantages of both approaches, assigning specific motivations to individual agents while leveraging global information to shape these motivations.

Beyond the literature on exploration, we discuss two other research lines that share similar techniques to those used in this work:

\textbf{Intrinsic Rewards for More than Exploration:}
In addition to leveraging intrinsic rewards for better exploration, the concept of intrinsic motivation has been applied to other aspects of MARL. For example, LIIR~\citep{LIIR} proposes utilizing intrinsic rewards to explicitly assign credits to different agents, resulting in an algorithm of enhanced performance. Other works design intrinsic rewards to incorporate preferences such as social influences~\citep{int_social}, social diversity~\citep{int_social_div}, and alignment\citep{ELIGN,int_consistency} into the learned policies.

\textbf{Credit Assignment:}
Credit assignment is a key challenge in MARL, referring to how to allocate global rewards to provide accurate feedback for individual agents~\citep{marl_survey}. Several value decomposition methods, such as VDN~\citep{VDN}, QMIX~\citep{QMIX}, and QTRAN~\citep{QTRAN}, have been proposed to implicitly assign credits among agents for discrete actions, and a later work LICA~\citep{LICA} tackles the same issue in the continuous action domain. COMA~\citep{COMA} takes a different approach and uses the counterfactual baseline to explicitly measure each agent's contribution. More recently, NA2Q~\citep{NA2Q} proposes an interpretable credit assignment framework by exploiting generalized additive models. Unlike these works that aim to solve the credit assignment challenge, our work focuses on avoiding extra complexity brought by non-stationary intrinsic rewards to the original credit assignment problem.

\section{Individual Contributions as Intrinsic Exploration Scaffolds}
In this work, we propose a novel approach of leveraging individual contributions as intrinsic scaffolds to enhance exploration in MARL. It aims to fully utilize privileged global information during centralized training while ensuring decentralized execution remains unaffected.

The following subsections will address three key questions:
1) \textbf{Why use individual contributions as intrinsic scaffolds?} \cref{sec: global_to_individual} examines the advantages of focusing on the contributions of individual agents over collective team efforts in enhancing exploration strategies within MARL.
2) \textbf{How to assess individual contributions?} \cref{sec: scaffolds_construction} describes how our methods quantify each agent's impact on global latent state transitions using Bayesian surprise.
3) \textbf{How are these scaffolds utilized effectively?}
\cref{sec: MARL_training} elaborates on how exploration and exploitation policies are optimized with distinct objectives and strategies to utilize these scaffolds effectively, thereby enhancing exploration without compromising the original training objectives or the decentralized execution strategy.

\subsection{From Global Intrinsic Rewards to Individual Contributions} \label{sec: global_to_individual}
Previous methods largely rely on formulating a global intrinsic reward, which is then added to extrinsic rewards to incentivize agents to explore. This strategy presents two notable drawbacks: Firstly, adding intrinsic rewards to the existing extrinsic rewards alters the original training objective. Intrinsic rewards, often learned and thus non-stationary throughout the training phase~\cite{curiosity_analysis}, will introduce additional non-stationarity into the training objective.

Secondly, like global extrinsic rewards, global intrinsic rewards require credit assignment among agents, a task that becomes more challenging with the non-stationary nature of intrinsic rewards. These complications can be effectively bypassed by directly providing agents with individual intrinsic motivations. In our method, this is achieved by utilizing privileged global information available only during the centralized training phase, thus addressing the issues of non-stationarity and complex credit assignment.

This is further verified by empirical ablation studies in \cref{sec: ablations}.

\subsection{Assessing Individual Contributions to Construct Intrinsic Scaffolds} \label{sec: scaffolds_construction}

\textbf{Bayesian Surprise to Characterize Individual Contributions:} In this subsection, we assess the individual contribution, denoted as $r^i_{t, \text{int}}$, of a specific action $u_t^i$ executed by agent $i$. The objective is to evaluate the impact of action $u_t^i$ on the global latent state transitions.

\begin{figure}[ht]
% \vskip 0.2in
\begin{center}
\centerline{\includegraphics[scale=1]{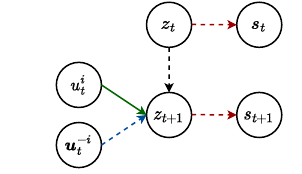}}
\caption{Dynamics model. The variable $z$ denotes the latent state. The solid line indicates the individual contributions of agent $i$'s actions, highlighted by the green arrow, signifying the primary focus of our measurement. Dashed lines represent other influences on state uncertainties, including the actions of other agents (blue arrow), which are excluded from agent $i$'s contribution assessment, and the environment's inherent stochasticity (red arrows), known as the noisy TV problem, which we aim to mitigate.\looseness=-1}
\label{fig:dynamics}
\end{center}
% \vskip -0.2in
\end{figure}

\begin{figure}[ht]
% \vskip -0.2in
\begin{center}
\centerline{\includegraphics[scale=1]{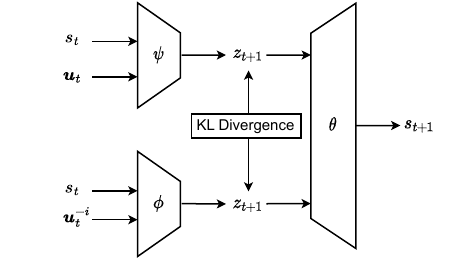}}
\caption{Modules for Bayesian surprise estimation. The structure resembles the CVAE structure with separate encoders and a shared decoder. The KL divergence between estimated priors is used as intrinsic contribution measurements. \looseness=-1}
\label{fig:CVAE}
\end{center}
% \vskip -0.4in
\end{figure}

\begin{figure*}[ht]
% \vskip 0.2in
\begin{center}
\centerline{\includegraphics[scale=1]{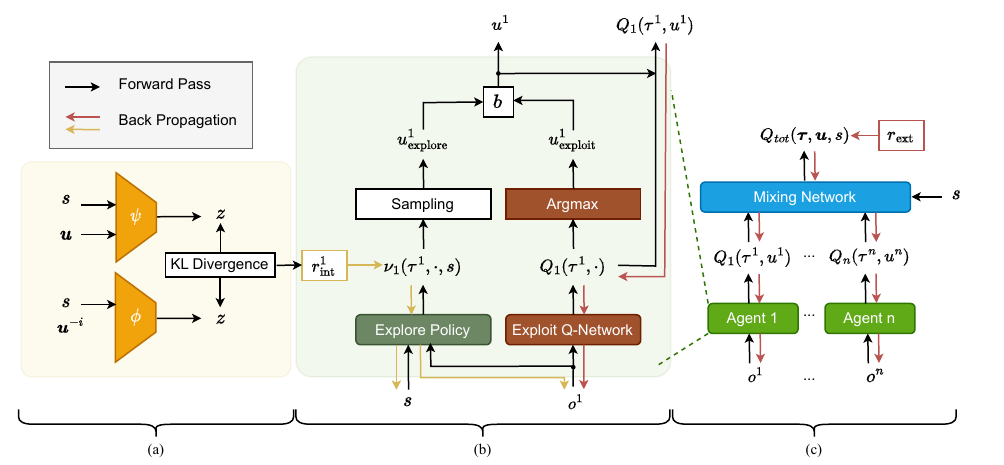}}
\caption{Network architecture of ICES. (a) Intrinsic exploration scaffolds. (b) Agent architecture. (c) Overall architecture. The \textcolor[HTML]{B85450}{red arrows} denote the gradient flows guided by the global extrinsic reward $r_\text{ext}$, and the \textcolor[HTML]{D6B656}{yellow arrows} denote the gradient flows guided by the individual scaffolds $r^i_\text{int}$. The training objectives for the exploration policy and exploitation policy are decoupled while both policies are combined for action selection during the training phase.}
\label{fig:arch}
\end{center}
\vskip -0.3in
\end{figure*}

To achieve this, we first demonstrate the environment dynamic model in \cref{fig:dynamics}, illustrating how state transition uncertainties are influenced by several factors. These include the impact of agent $i$'s action, represented by a mutual information term $r^i_{t, \text{int}} = I(z_{t+1}; u_t^i | s_t, \boldsymbol{u}_t^{-i})$; the influence of other agents' actions, denoted by $I(z_{t+1}; \boldsymbol{u}_t^{-i} | s_t, u_t^i)$; and the inherent environmental uncertainties, expressed as the entropy $H(s_t | z_t)$, known as the noisy TV problem~\citep{NoisyTV}. Our focus is primarily on the individual contribution $r^i_{t, \text{int}}$, which necessitates a specific measurement method to effectively distinguish the contribution of agent $i$'s action $u_t^i$ and mitigate potential misattributions among agents and the effects of noisy TV. Consequently, we employ the Bayesian surprise as the measurement method. Following previous works~\citep{Bayesian_surprise, Bayesian_curiosity_latent}, we express the contribution $r^i_{t, \text{int}}$ as the mutual information between the latent variable $z_{t+1}$ and the action $u_t^i$, which is given as 
\begin{align}
    r^i_{t, \text{int}} &= I(z_{t+1}; u_t^i | s_t, \boldsymbol{u}_t^{-i}) \nonumber\\
     &= D_{\text{KL}} \left[ p(z_{t+1}|s_t, \boldsymbol{u}_t) \parallel p(z_{t+1}|s_t, \boldsymbol{u}_t^{-i})  \right].
     \label{eq: individual_scaffolds}
\end{align}
\vskip -0.2in
This term captures the discrepancy between the actual and counterfactual latent state distributions from the perspective of an individual agent $i$. In later sections, we omit the subscript $t$ where contextually clear, referring to $r^i_{t, \text{int}}$ simply as $r^i_{\text{int}}$. \looseness=-1

\textbf{CVAE to Estimate the Bayesian Surprise:} For robust estimation of individual contributions, it is essential to identify a latent space for $z_t$ that is both compact and informative, capable of reconstructing the original state space. To achieve this, we resort to CVAE owing to its ability to induce explicit prior distributions and perform probabilistic inference, a necessity in environments with inherent stochasticity, such as GRF~\citep{GRF}. 

We aim to estimate two specific priors: $p_\psi(z_{t+1}|s_t, \boldsymbol{u}_t)$ and $ p_\phi(z_{t+1}|s_t, \boldsymbol{u}_t^{-i})$. However, learning these two priors independently will not yield satisfactory results, as shown later in our ablation studies (~\cref{fig:ablation_others}). This is due to the potential misalignment of the latent spaces created by each independently trained priors, rendering the KL-divergence measure less effective. Thus, to align the latent spaces, we use two separate encoders for these estimations while utilizing a shared decoder for reconstruction. \looseness=-1
% , following a similar methodology to previous work~\citep{ME_VAE}.

The CVAE's architecture, depicted in \cref{fig:CVAE}, includes the following components:
\begin{equation*}
    \begin{split}
        & \textrm{Prior Encoders:} \\
        \\
        & \textrm{Reconstruction Decoder:} \\
        & \textrm{Latent Posteriors:} \\
        \\
    \end{split}
    \quad
    \begin{split}
%        &  p(s_{t}|o_t) \\
        & p_\psi(z_{t+1} | s_t, \boldsymbol{u}_t),  \\
        & p_\phi(z_{t+1}|s_t, \boldsymbol{u}_t^{-i}), \\
        &  p_\theta(s_{t+1}|z_{t+1}), \\
        & q_\psi(z_{t+1}|s_t, \boldsymbol{u}_t, s_{t+1} ), \\
        & q_\phi(z_{t+1}|s_t, \boldsymbol{u}_t^{-i}, s_{t+1} ). \\
    \end{split}
\end{equation*}

\textbf{Training Objective for Scaffolds:} The training objective of the above modules is to maximize the variational lower bound of the conditional log-likelihood~\citep{CVAE}, formalized as: \looseness=-1
\begin{align}
    \mathcal{J}(\psi &, \phi , \theta)\! = \!-\! D_{\text{KL}} \left[ q_\psi(z_{t+1}|s_t, \boldsymbol{u}_t, s_{t+1} ) \!\!\parallel\! p_\psi(z_{t+1} | s_t, \boldsymbol{u}_t)   \right] \nonumber\\
    &- D_{\text{KL}} \left[ q_\phi(z_{t+1}|s_t, \boldsymbol{u}_t^{-i}, s_{t+1} ) \parallel p_\phi(z_{t+1}|s_t, \boldsymbol{u}_t^{-i}) \right] \nonumber \\
    &+ \mathbb{E}_{z \sim q_\psi} \left[\log p_\theta(s_{t+1}|z) \right] + \mathbb{E}_{z \sim q_\phi} \left[\log p_\theta(s_{t+1}|z) \right]. \nonumber\\
    \label{eq: obj_scaffolds}
\end{align}

\subsection{Decoupling Exploration and Exploitation Policies to Utilize Intrinsic Scaffolds} \label{sec: MARL_training}

In the ICES framework, we retain the training objective of learning the target (exploitation) policy $\boldsymbol{\pi}$, aiming at maximizing the cumulative reward $G_t = \sum_t \gamma^t r^t$. Concurrently, we adjust the behavior policy $\boldsymbol{b} = \{b_i\}_{i=1}^n$ to enhance exploration. Unlike the classical $\epsilon$-greedy method adopted by most of the off-policy works, where actions are uniformly sampled if not following the target policy, ICES agents prioritize actions that significantly contribute to state transitions, as identified by $r^i_\text{int}$. The overall architecture is depicted in \cref{fig:arch}, with different gradient flows denoted by \textcolor[HTML]{B85450}{red} and \textcolor[HTML]{D6B656}{yellow} arrows for global extrinsic rewards and individual scaffolds, respectively.

We denote the target policy as $\boldsymbol{\pi} = \{\pi_i\}_{i=1}^n$, the exploration policy as $\{\nu_i\}_{i=1}^n$ and the behavior policy derived from the above two policies as $\boldsymbol{b} = \{b_i\}_{i=1}^n$.

\textbf{Combining the Exploration and Exploitation Policies for Behavior Policies:}
As shown in part (b) of \cref{fig:arch}, action selection involves both the exploration policy $\nu_i$ and the exploitation policy $\pi_i$. The behavior policy $b_i$ is determined as follows:
\begin{align}
    u^i &\sim b_i\left(u^i_\text{explore}, u^i_\text{exploit}\right) \nonumber\\
    &= 
    \begin{cases}
    u^i_\text{explore} &\text{with probability } \alpha \\
    u^i_\text{exploit} &\text{with probability } 1 - \alpha
    \end{cases}
    ,
    \label{eq: bahavior_policy}
\end{align}
where $\alpha$ is a hyperparameter balancing exploration and exploitation during training. The exploration and exploitation actions are given as:
\begin{align}
    u^i_\text{explore} &\sim \nu_i(\tau^i, u, s), \label{eq: exploration_policy}\\
    u^i_\text{exploit} &= \argmax_u Q_i(\tau^i, u),
\end{align}
with $Q_i(\tau^i, \cdot)$ representing the local Q-value function for exploitation. \looseness=-1

\textbf{Optimization Objectives:} The optimization objective for the exploitation policy, parameterized by $\zeta$, is to minimize the TD-error loss:
\begin{align}
    \mathcal{L}(\zeta) = \mathbb{E}_{(\boldsymbol{\tau}_t, \boldsymbol{u}_t, s_t, r_\text{ext}, s_{t+1}) \sim \mathcal{D}} \left[\left(y^{tot} - Q_{tot}(\boldsymbol{\tau}_t, \boldsymbol{u}_t, s_t; \zeta) \right)^2\right],
    \label{eq: obj_ext}
\end{align}
where $y^{tot} = r_\text{ext} + \gamma \max_{\boldsymbol{u}} Q_{tot}(\boldsymbol{\tau}_{t+1}, \boldsymbol{u}, s_{t+1}; \zeta^-)$ and $\zeta^-$ are the parameters of a target network as in DQN.

The optimization objective for the exploration policy, parameterized by $\xi$, is to maximize the average individual intrinsic scaffolds (not the episodic return objective) and exploration policy entropy: \looseness=-1
\begin{align}
    \mathcal{J}_i(\xi) = \mathbb{E}_{\nu_i}[r^i_\text{int}] + \beta \mathcal{H}(\cdot |\tau^i, s),
    \label{eq: obj_int}
\end{align}
where $\beta$ is a hyperparameter to control the regularization weight for entropy maximization and $\mathcal{H}(\cdot |\tau^i, s) = -\mathbb{E}_{\nu_i(\xi)} \ln{\nu_i(\cdot |\tau^i, s; \xi)}$ is the entropy of policy $\nu_i$ at local-global observation pair $(\tau^i, s)$.

This approach ensures that while the training of the exploitation network remains centralized and retained, the exploration network benefits from decentralized training guided by individual intrinsic scaffolds.  This strategy circumvents the challenges in intrinsic credit assignment. Moreover, since exploration policies are employed only during training, they can utilize privileged information, such as the global observation $s$, for more informed decision-making.

\textbf{REINFORCE with Baseline for Exploration Policy Training:}
By decoupling the exploration and exploitation, we can employ distinct RL algorithms to update each policy, leveraging their respective strengths. For the exploitation policy update (denoted by the \textcolor[HTML]{B85450}{red arrows} in \cref{fig:arch}), we follow the previous work and use the DQN update with value decomposition methods like QMIX~\citep{QMIX} or QPLEX~\citep{QPLEX}. 

For the exploration policy, whose gradient is denoted by the \textcolor[HTML]{D6B656}{yellow arrows} in \cref{fig:arch}, we prefer stochastic policies over deterministic ones for more diverse behaviors. Thus, we adopt a policy-based reinforcement learning algorithm with entropy regularization with the objective function given in ~\cref{eq: obj_int}. To stabilize training, we introduce a value function $V(\tau^i, s; \eta)$ as a baseline. With the policy gradient theorem (details elaborated in \cref{sec: appendix_exploration_gradient}), we arrive at 
\vskip -0.2in
\begin{align}
    \nabla_\xi \mathcal{J}_i(\xi) = \mathbb{E}_{\nu_i(\xi), (\tau_i, s) \sim \mathcal{D}} \left[ A \cdot \nabla_\xi \ln \nu(\cdot |\tau^i, s; \xi)   \right], 
    \label{eq: exploration_gradient}
\end{align}
where $A = r^i_\text{int} - V(\tau^i, s; \eta) - \beta $ is the advantage function and $V_\eta(\tau^i, s)$ is updated by minimizing 
\begin{align}
    \mathcal{L}(\eta) = \mathbb{E}_{\nu_i(\xi), (\tau_i, s) \sim \mathcal{D}} \left[ \left(r^i_\text{int} - V(\tau^i, s; \eta) \right)^2\right].
    \label{eq: obj_int_v}
\end{align}

% \vskip -2in
\subsection{Overall ICES Training Algorithm}\label{sec: algo_pseudo_codes}

We summarize the overall training procedure for ICES in \cref{algo: ICES_training}. In particular, we train a scaffolds network (updated by \cref{algo: TrainScaffolds} in \cref{sec: appendix_training_details}) with parameters $\psi, \phi, \theta$ and two policy networks (updated by \cref{algo: TrainPolicies} in \cref{sec: appendix_training_details}), including an exploration network parametrized by $\xi, \eta$ and an exploitation network parameterized by $\zeta$. 
% We update the scaffolds network and policy networks alternatively during training stage
We utilize the scaffolds network to provide guidance for exploration network updates, and we utilize the exploration network to influence the action selection processes, consequently influencing the learning process of exploitation networks. Among the above networks, only the exploitation network will be used for execution. 

\begin{algorithm}[ht]
\caption{Training Procedure of ICES}
\begin{algorithmic}[1]
\STATE {\bfseries Init:} Scaffolds parameters $\psi, \phi, \theta$
\STATE {\bfseries Init:} Exploration networks parameters $\xi, \eta$
\STATE {\bfseries Init:} Exploitation networks parameters $\zeta$
% \STATE {\bfseries Init:} Replay buffer $\mathcal{D} = \emptyset$, $\text{step} = 0$
% \STATE {\bfseries Init:} Target parameters $\theta^- =\theta$
\STATE {\bfseries Init:} $\mathcal{D} = \emptyset$, $\text{step} = 0$, $\theta^- =\theta$
\WHILE{$\text{step} < \text{step}_\text{max}$}
    \STATE $t=0$. Reset the environment. 
    
    \FOR{$t = 1, 2, ..., \text{episode\_limit}$} 
        \FOR{$i = 1, 2, ..., n$}
            \STATE Select actions $u_t^i \sim b_i$
            \hfill \COMMENT{$\triangleright$ \cref{eq: bahavior_policy}}
            % \hfill \COMMENT{$\triangleright$ Collect experience. (\cref{sec: MARL_training})}
        \ENDFOR
        % \STATE Interact with the environment. $(s_{t+1}, \boldsymbol{o}_{t+1}, r_{t, \text{ext}}) = \text{env}.\text{step}(\boldsymbol{u}_t)$.
        \STATE $(s_{t+1}, \boldsymbol{o}_{t+1}, r_{t, \text{ext}}) = \text{env}.\text{step}(\boldsymbol{u}_t)$
        % \STATE Save the transition tuple $\mathcal{D} = \mathcal{D} \cup (s_t, \boldsymbol{o}_t, \boldsymbol{u}_t, r_{t, \text{ext}}, s_{t+1}, \boldsymbol{o}_{t+1})$
        \STATE $\mathcal{D} = \mathcal{D} \cup (s_t, \boldsymbol{o}_t, \boldsymbol{u}_t, r_{t, \text{ext}}, s_{t+1}, \boldsymbol{o}_{t+1})$
    \ENDFOR
    \IF{$\text{step} \mod \text{train\_interval} == 0$} 
        \STATE $\xi, \eta, \zeta \leftarrow \text{TrainPolicies}(\psi, \phi, \xi, \eta, \zeta, \mathcal{D})$
        \\\hfill \COMMENT{$\triangleright$ \cref{algo: TrainPolicies} }
        \STATE $\psi, \phi, \theta \leftarrow \text{TrainScaffolds}(\psi, \phi, \theta, \mathcal{D})$
        \\\hfill \COMMENT{$\triangleright$ \cref{algo: TrainScaffolds} }
    \ENDIF
    \IF{$\text{step} \mod \text{target\_update\_interval} == 0$}
        \STATE $\theta^- = \theta$
    \ENDIF
\ENDWHILE
\STATE {\bfseries Output:} Exploitation networks parameters $\zeta$
% \RETURN{} Parameters for exploitation networks $\zeta$
\end{algorithmic}
\label{algo: ICES_training}
\end{algorithm}
% \vskip -0.2in

\begin{figure*}[t]
\centering
% \subfigure[]{
%         \includegraphics[width=0.9\textwidth]{figs/results/legend_1226.pdf}
%     }
\raisebox{-\height}{\includegraphics[width=0.95\textwidth]{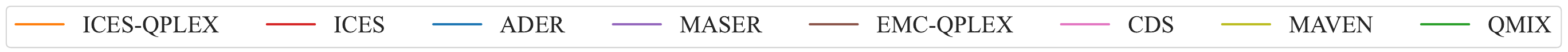}}
\subfigure[3\_vs\_1\_with\_keeper]{
        \includegraphics[width=0.23\textwidth]{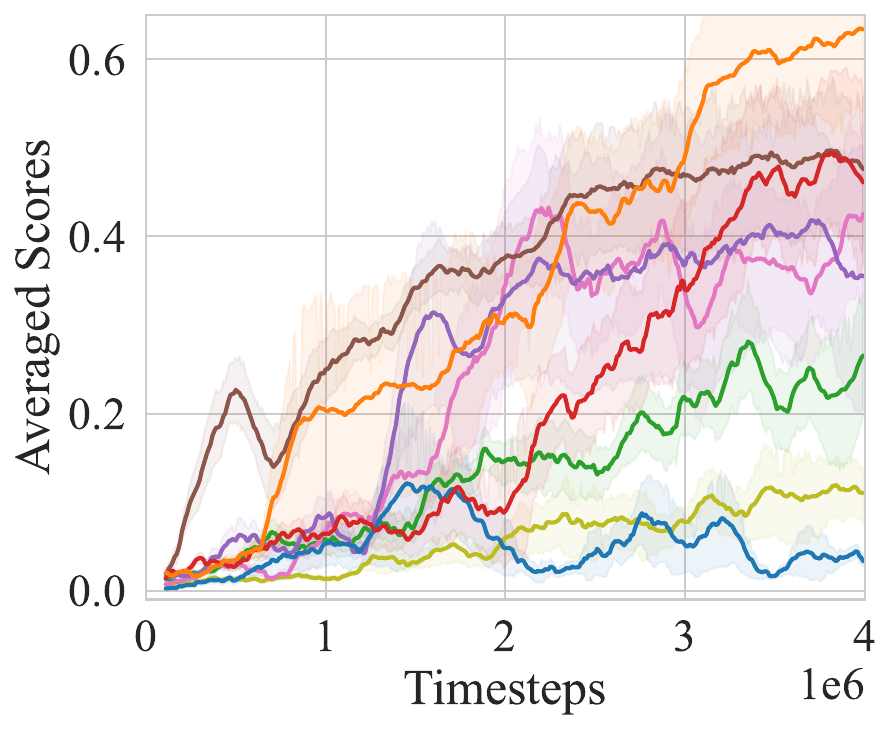}
        \label{fig:3v1}
    }
\subfigure[corner]{
        \includegraphics[width=0.23\textwidth]{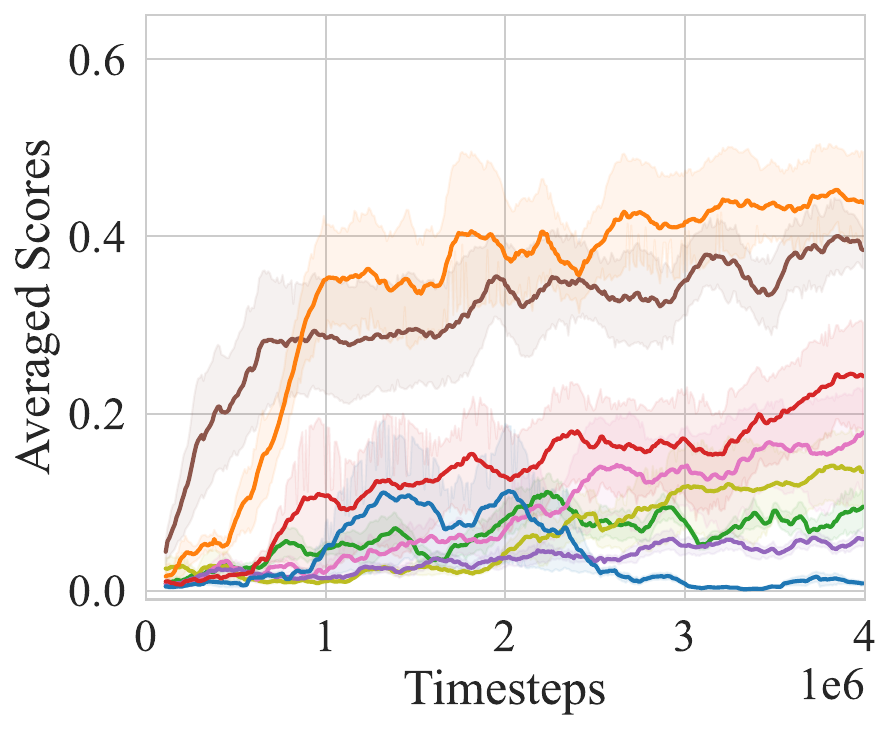}
        \label{fig:corner}
    }
\subfigure[counterattack\_hard]{
        \includegraphics[width=0.23\textwidth]{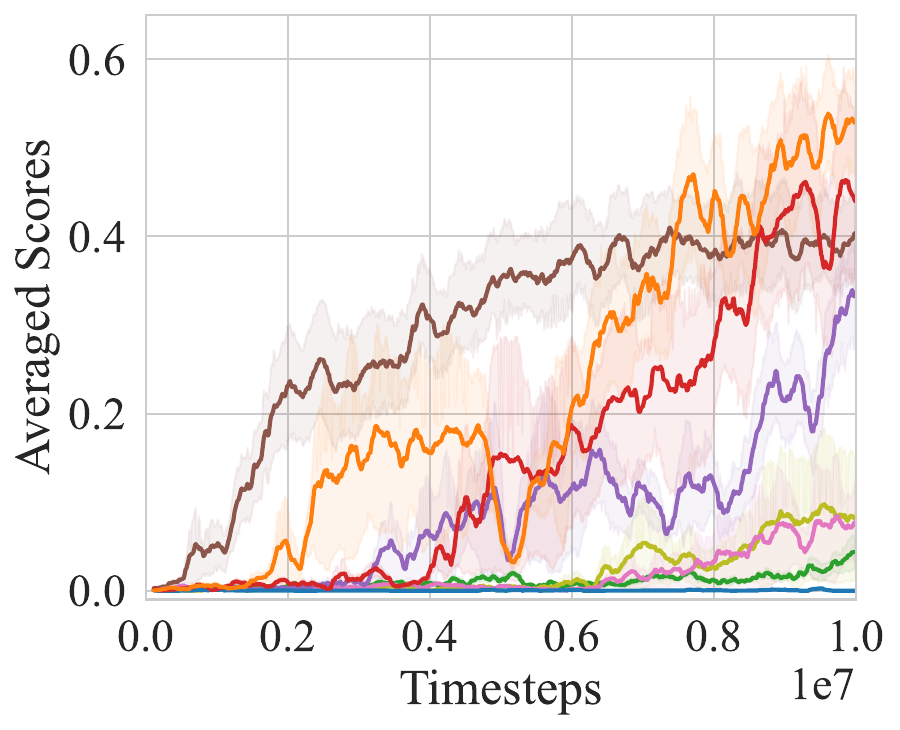}
        \label{fig:cah}
    }
\subfigure[3m]{
        \includegraphics[width=0.23\textwidth]{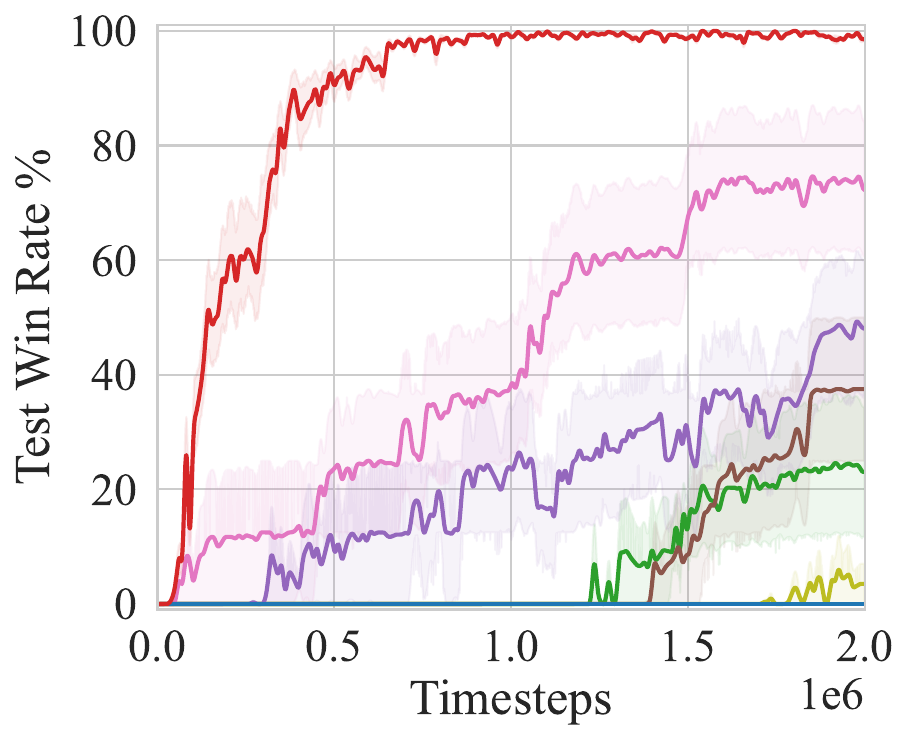}
        \label{fig:3m}
    }
\subfigure[8m]{
        \includegraphics[width=0.23\textwidth]{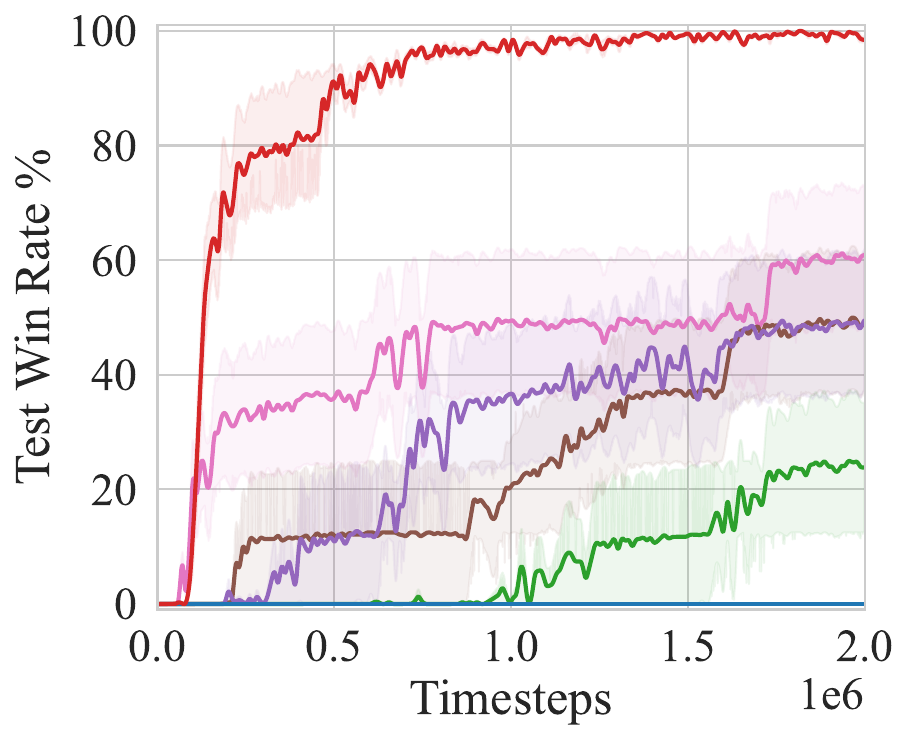}
        \label{fig:8m}
    }
\subfigure[2s3z]{
        \includegraphics[width=0.23\textwidth]{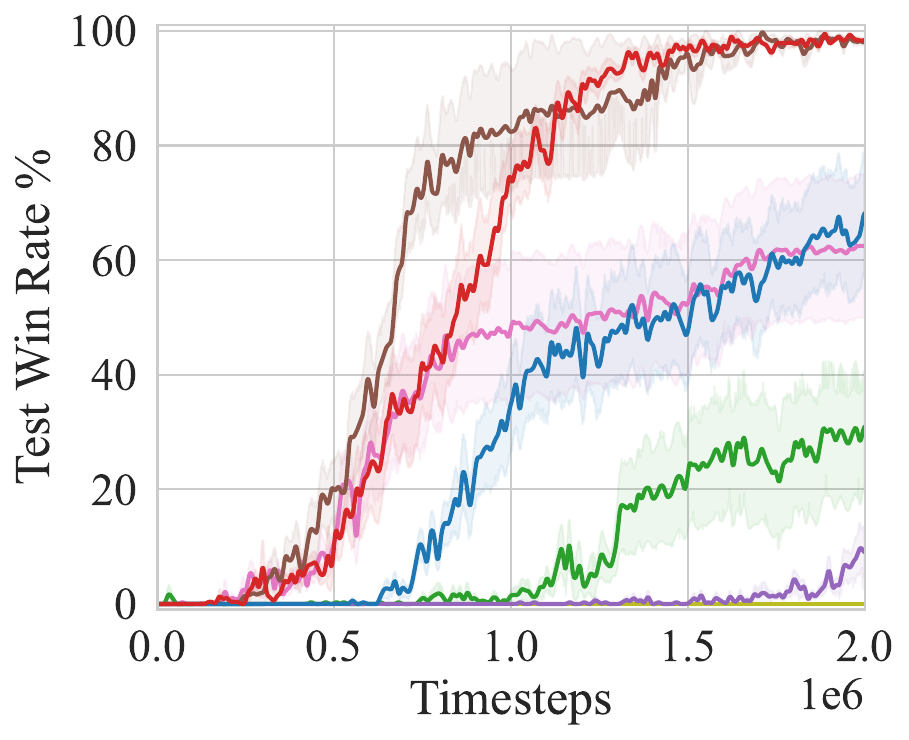}
        \label{fig:2s3z}
    }
\subfigure[2s\_vs\_1sc]{
        \includegraphics[width=0.23\textwidth]{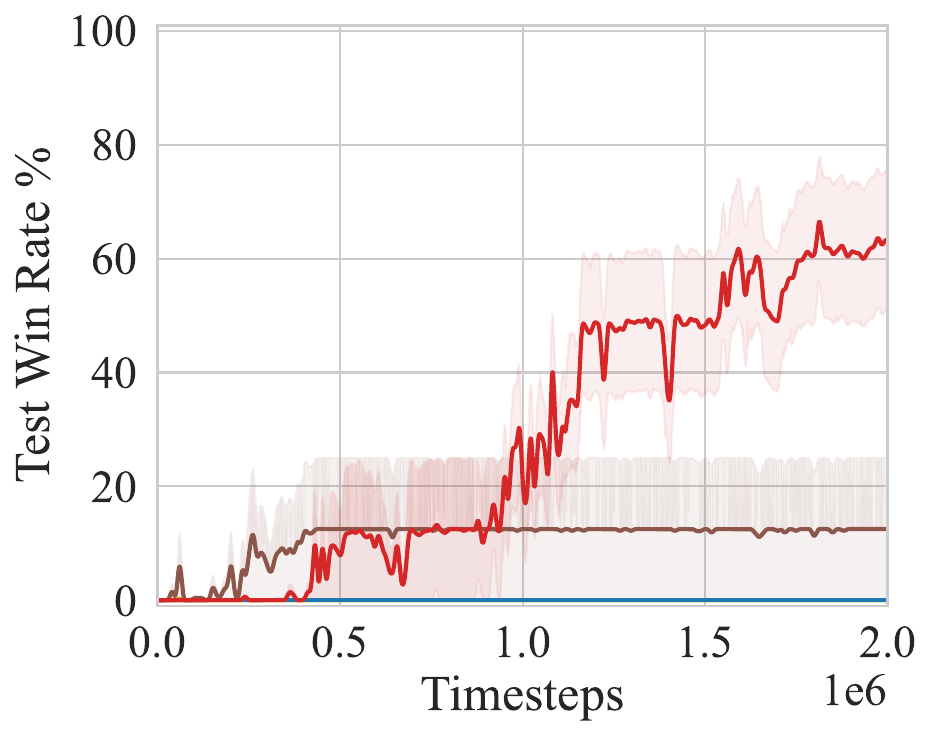}
        \label{fig:2s_vs_1sc}
    }
\subfigure[5m\_vs\_6m]{
        \includegraphics[width=0.23\textwidth]{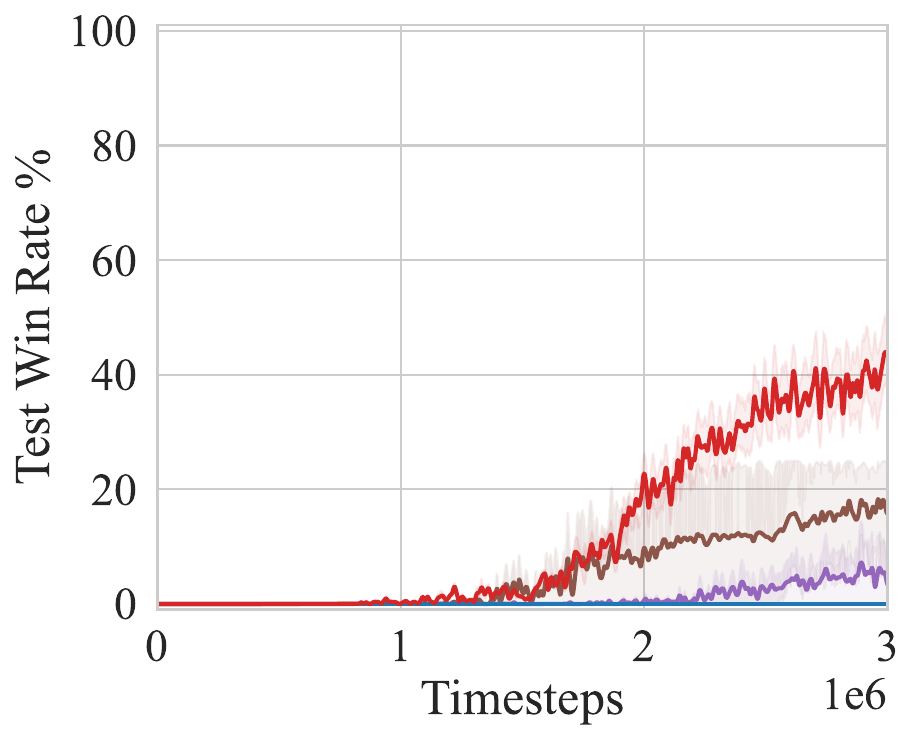}
        \label{fig:5m_vs_6m}
    }

\caption{Performance comparison with baselines on GRF and SMAC benchmarks in sparse reward settings.}
\label{fig:results_grf_smac}
% \vskip -0.3in
% \vskip -0.2in
\end{figure*}

\section{Experiments}
In this section, we evaluate ICES on two multi-agent benchmark tasks: GRF and SMAC. Experiments in the GRF domain are averaged over eight random seeds and experiments in the SMAC domain are averaged over five random seeds. The shaded areas represent 50\% confidence intervals. Detailed descriptions of network architectures and training hyperparameters are available in \cref{sec: appendix_implementation_details}. Further experimental results are provided in \cref{sec: appendix_more_exp}. \looseness=-1

% \vskip -2in
\subsection{Settings}
\textbf{Benchmarks: } In this work, we test ICES and baselines on widely used benchmarks of GRF and SMAC~\footnote{We use SC2.4.10 with difficulty of 7. Note that performance is not always comparable across versions.} in sparse reward settings, with details elaborated in \cref{sec: appendix_env}.

\textbf{Baselines: } We implement our proposed ICES on top of QMIX~\citep{QMIX}. For the GRF environment, we add a curve combining ICES with QPLEX~\citep{QPLEX} and denote the results as ICES-QPLEX. We compare ICES with six state-of-arts baselines: ADER~\citep{ADER}, MASER~\citep{MASER}, EMC~\citep{EMC} (built upon QPLEX, denoted as EMC-QPLEX), CDS~\citep{CDS}, MAVEN~\citep{MAVEN} and QMIX~\citep{QMIX}. Wherever possible, we utilize the official implementations of these baselines from their respective papers; in cases where the implementation is not available, we closely follow the descriptions provided in the papers and implement them on top of QMIX.

\subsection{Benchmark results on GRF and SMAC}
We present the comparative performance of ICES and various baselines in \cref{fig:results_grf_smac}. Overall, ICES demonstrates superior performance over baselines constructed on QMIX. When integrated with QPLEX (ICES-QPLEX), it surpasses baselines built on QPLEX. This showcases that ICES is able to foster effective exploration in MARL training, without tampering with the original training objective (discussed in \cref{sec: MARL_training}), thus leading to a fast convergence and enhanced final performance. Challenges in GRF, including environmental stochasticity and the need for agent collaboration, are adeptly addressed by ICES. In particular, ICES filters out environmental noise using Bayesian surprise and promotes team coordination by constructing scaffolds based on global state transitions. (as discussed in \cref{sec: scaffolds_construction}.) It is also worth mentioning that, among the baselines, EMC-QPLEX also shows notable exploration capabilities, particularly in the early stages of training. This suggests that utilizing episodic memory, as EMC-QPLEX does, could be a beneficial approach to improve sample efficiency, albeit different from our focus on generating more informative exploration experiences. \looseness=-1

 For SMAC tasks, ICES also demonstrates strong performance compared with baselines, where most of the baselines require more training budget to find the winning strategy. For example, in scenario \texttt{2s\_vs\_1sc}, ICES starts to observe winning episodes while baselines have not after 2M timesteps. In SMAC, the exploration challenges mainly arise from the large state space, which ICES addresses by computing intrinsic scaffolds within a compact latent space, rather than the extensive original state space.
 
\begin{figure}[t]
\centering
\raisebox{-\height}{\includegraphics[width=0.45\textwidth]{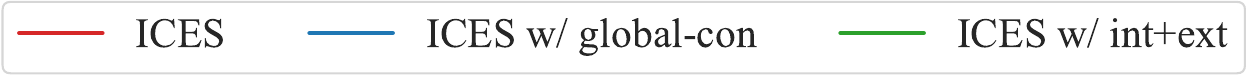}}
\subfigure[3\_vs\_1\_with\_keeper]{
        \includegraphics[width=0.22\textwidth]{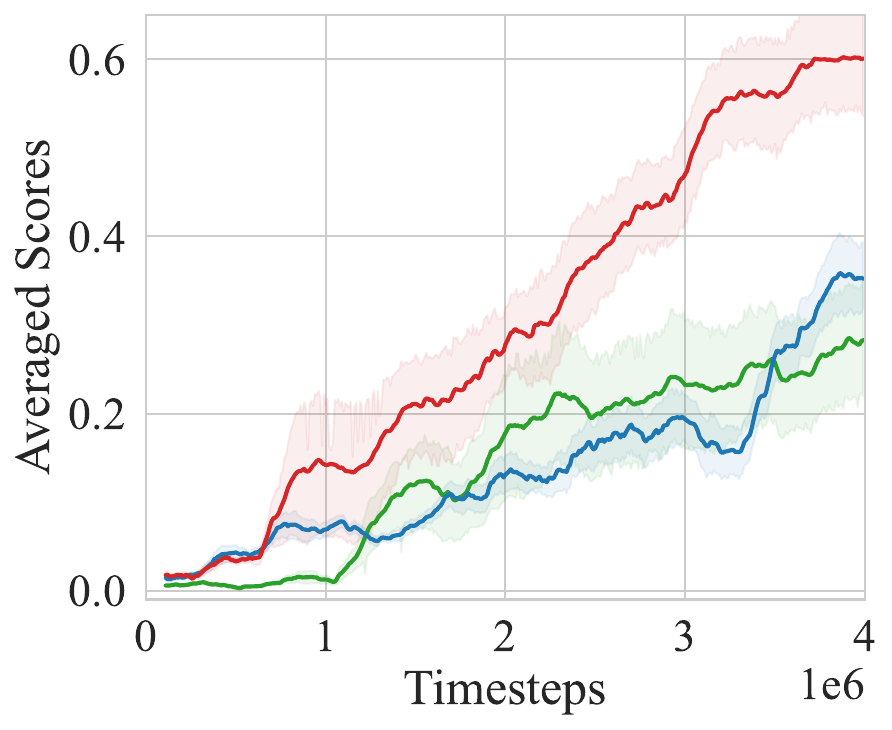}
    }
\subfigure[2s\_vs\_1sc]{
        \includegraphics[width=0.22\textwidth]{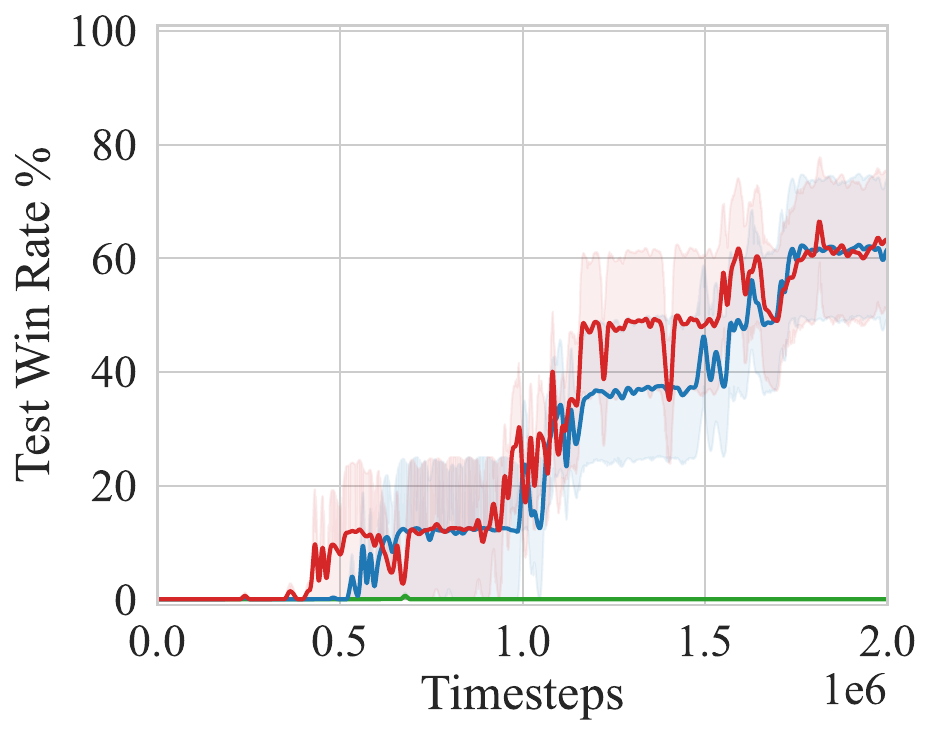}
    }
\caption{Ablations on individual contributions.}
\label{fig:ablation_ind}
% \vskip -0.2in
\end{figure}

\begin{figure}[t]
\centering
\raisebox{-\height}{\includegraphics[width=0.4\textwidth]{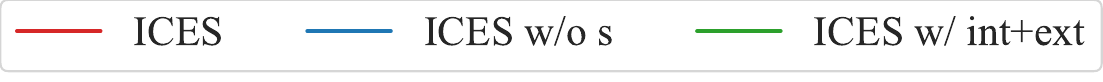}}
\subfigure[3\_vs\_1\_with\_keeper]{
        \includegraphics[width=0.22\textwidth]{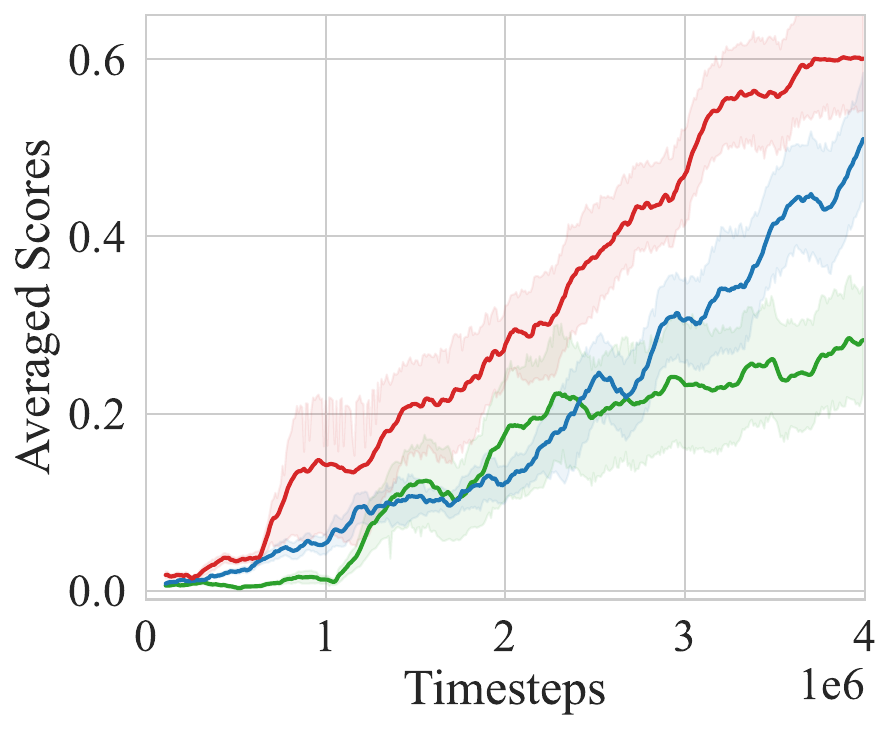}
    }
\subfigure[2s\_vs\_1sc]{
        \includegraphics[width=0.22\textwidth]{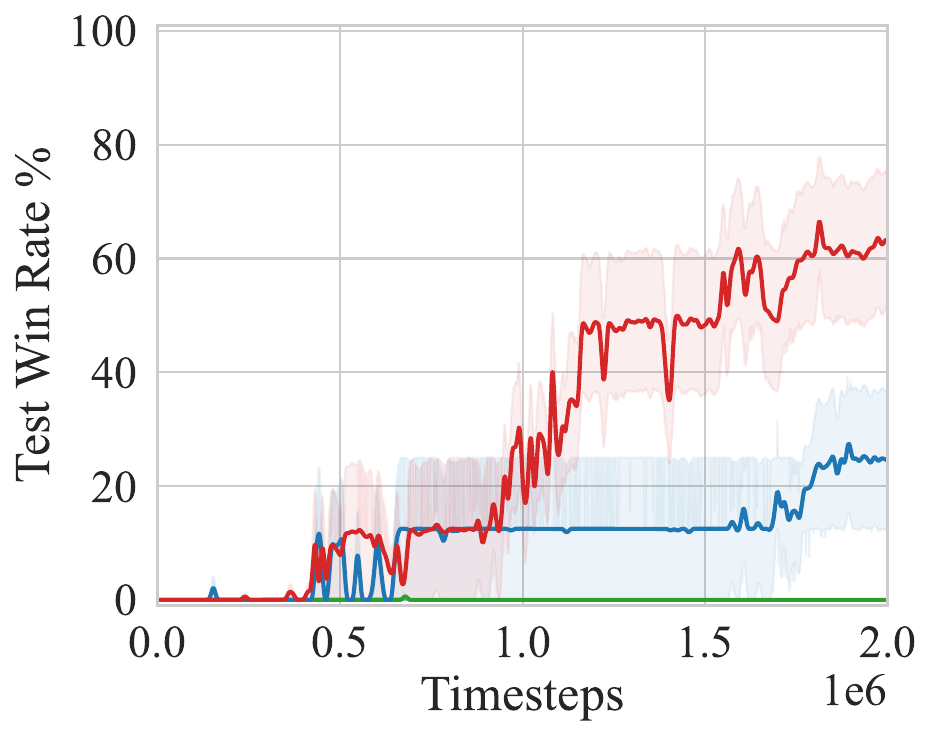}
    }
\caption{Ablations on decoupling exploration and exploitation policies.}
\label{fig:ablation_entangled}
% \vskip -0.2in
\end{figure}

\begin{figure}[t]
\centering
\raisebox{-\height}{\includegraphics[width=0.45\textwidth]{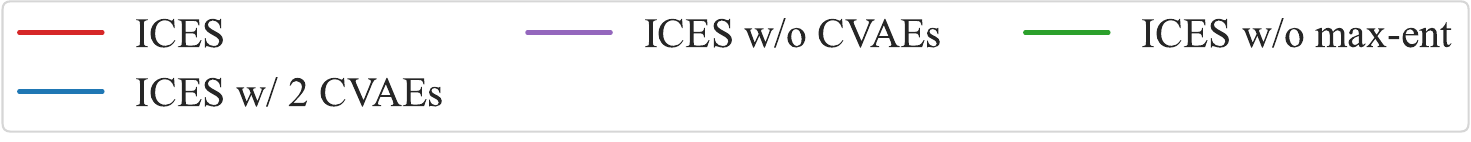}}
\subfigure[3\_vs\_1\_with\_keeper]{
        \includegraphics[width=0.22\textwidth]{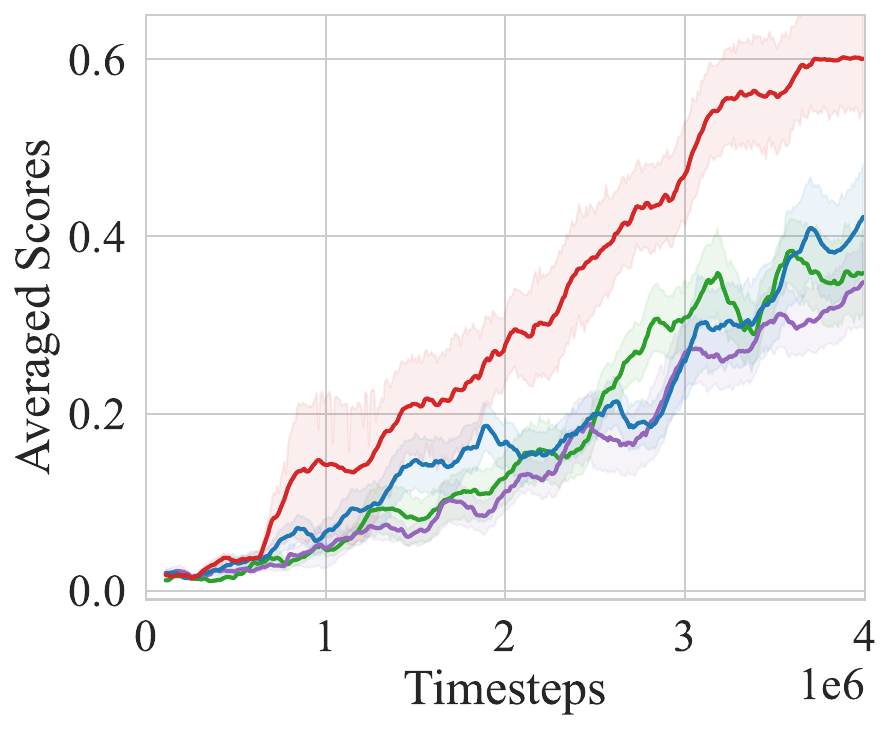}
    }
\subfigure[2s\_vs\_1sc]{
        \includegraphics[width=0.22\textwidth]{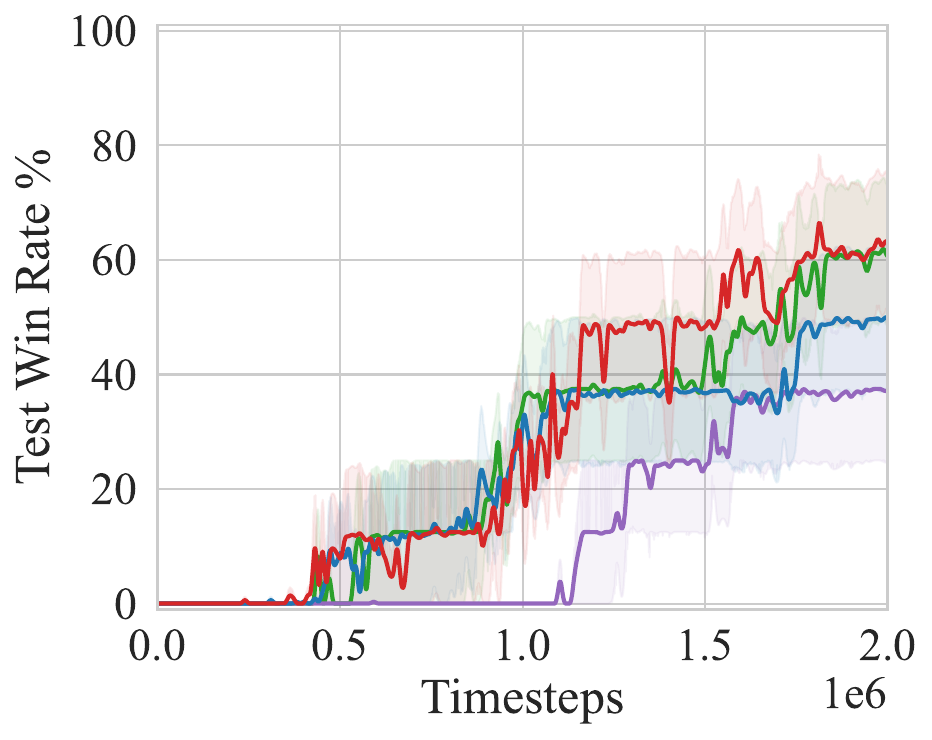}
    }
\caption{Ablations on other design choices.}
\label{fig:ablation_others}
% \vskip -0.3in
\end{figure}

\begin{figure}[t]
\centering

\subfigure[Analysis on $\alpha$]{
        \includegraphics[width=0.22\textwidth]{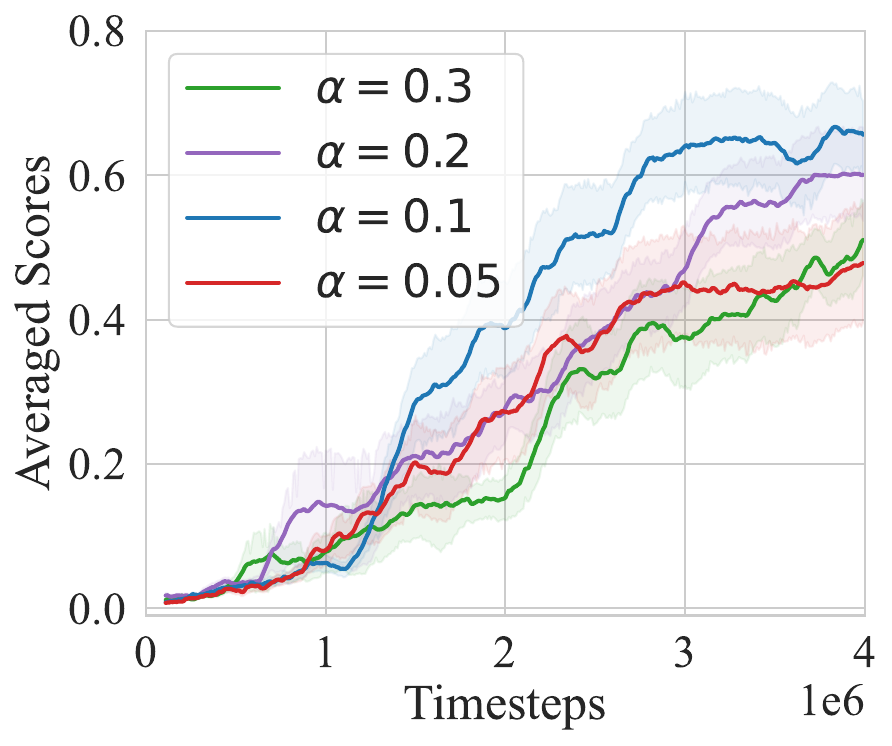}
    }
\subfigure[Analysis on $\beta$]{
        \includegraphics[width=0.22\textwidth]{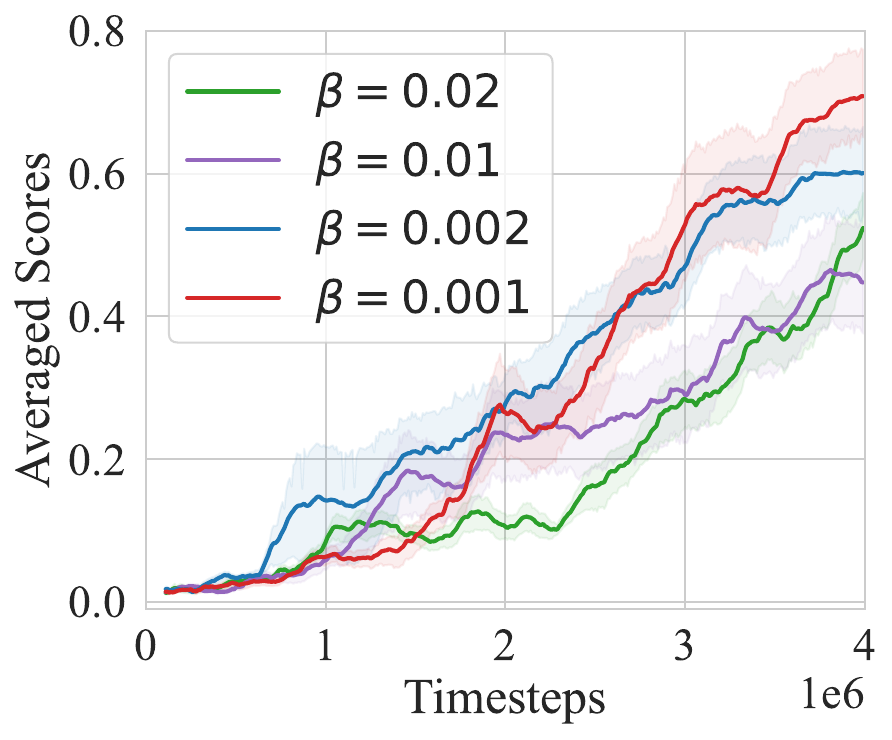}
    }
\caption{Hyperparameter analysis on the 3\_vs\_1\_with\_keeper scenario.}
\label{fig:hyper_ana}
\vskip -0.3in
\end{figure}

\begin{figure*}[t]
% \raisebox{-\height}
{\includegraphics[width=0.24\textwidth]{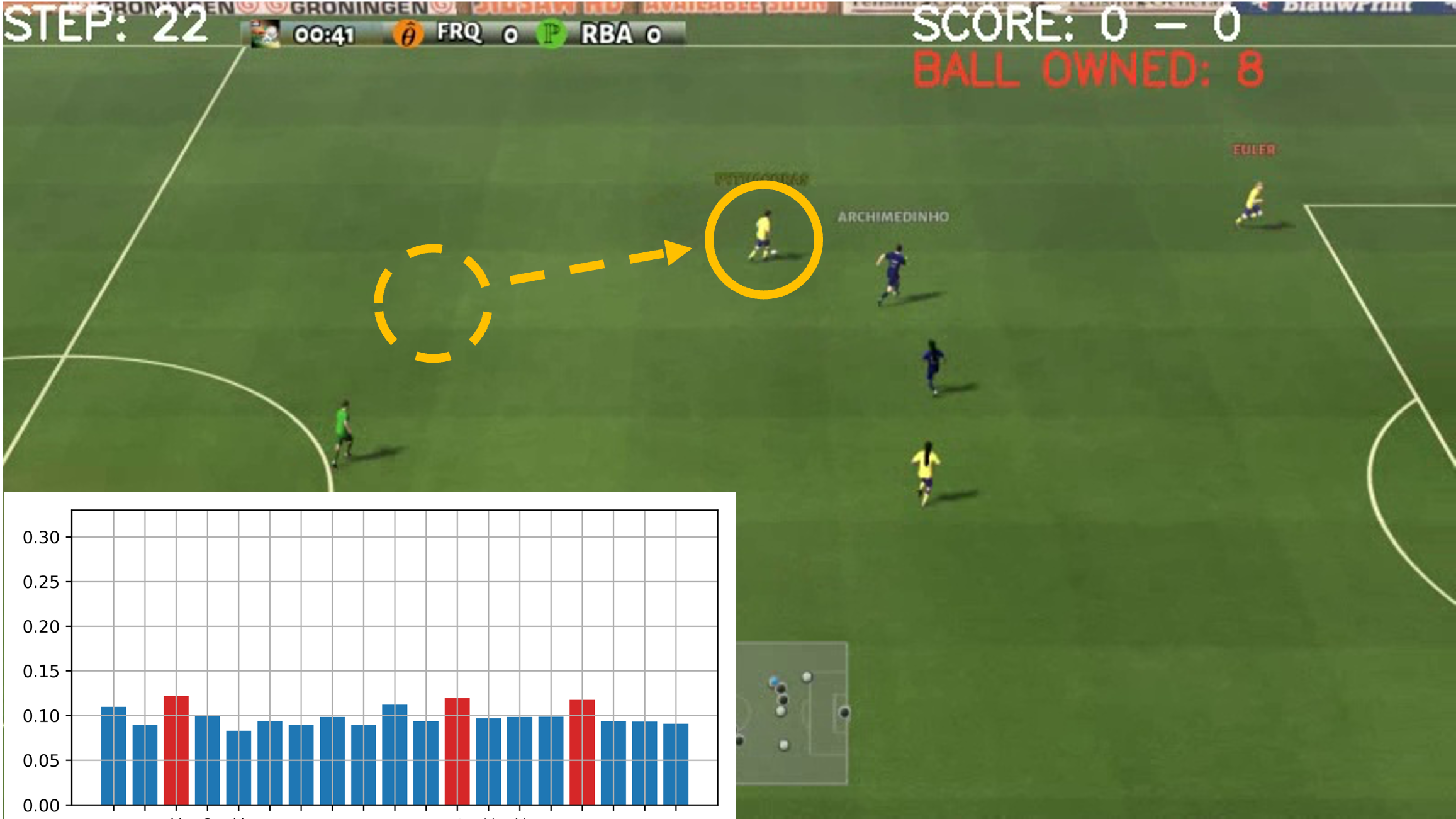}}
{\includegraphics[width=0.24\textwidth]{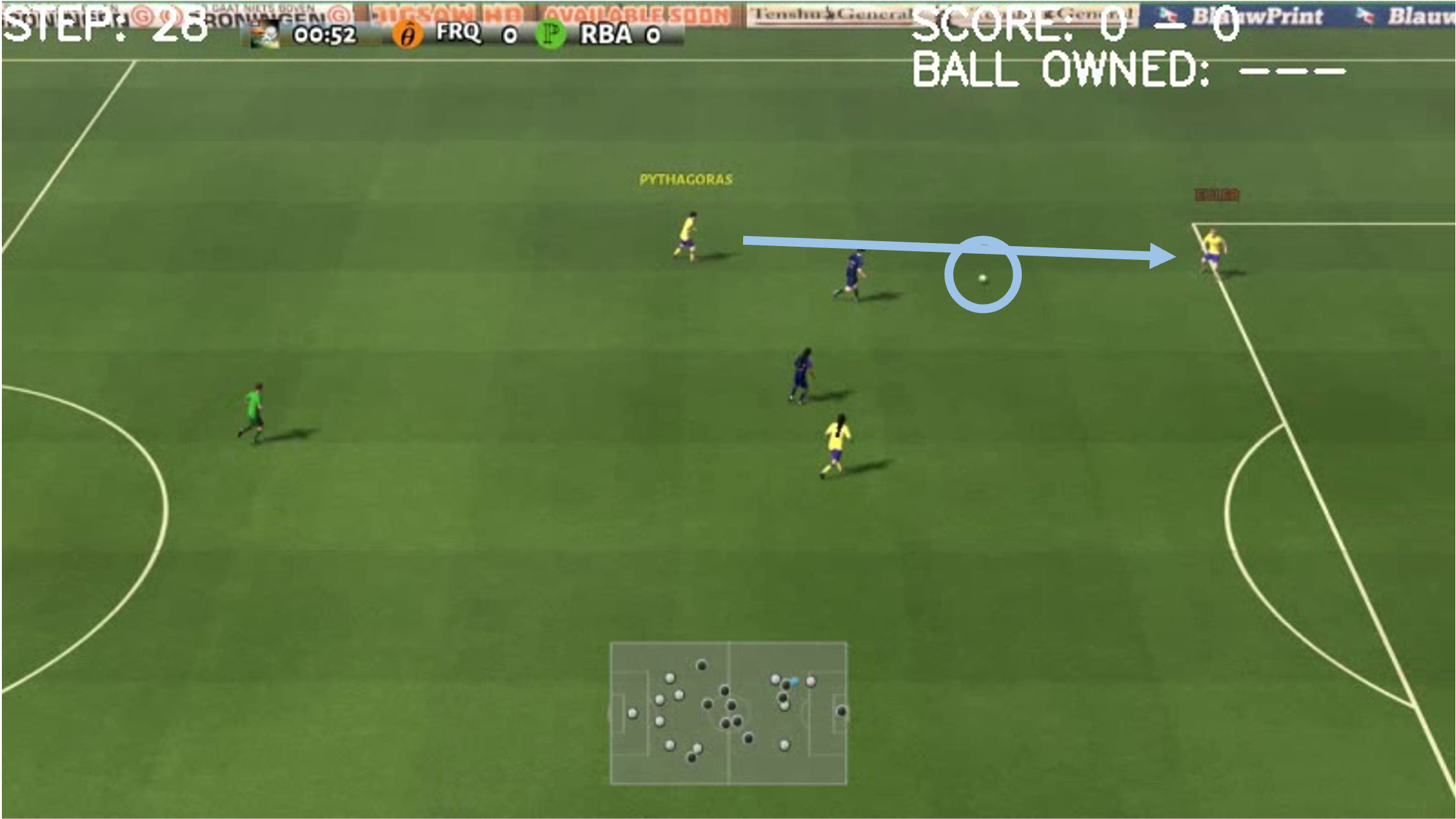}}
{\includegraphics[width=0.24\textwidth]{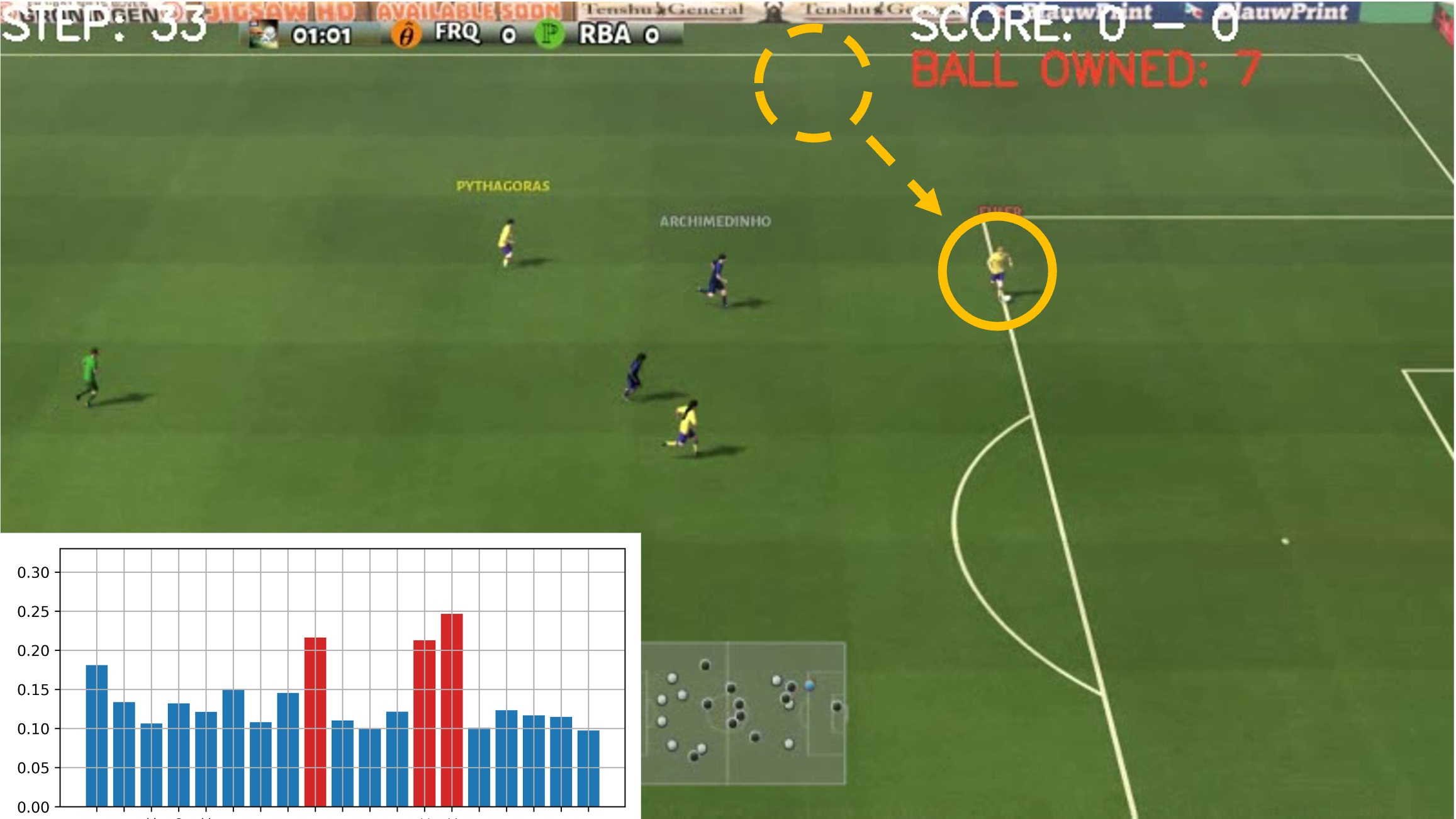}}
{\includegraphics[width=0.24\textwidth]{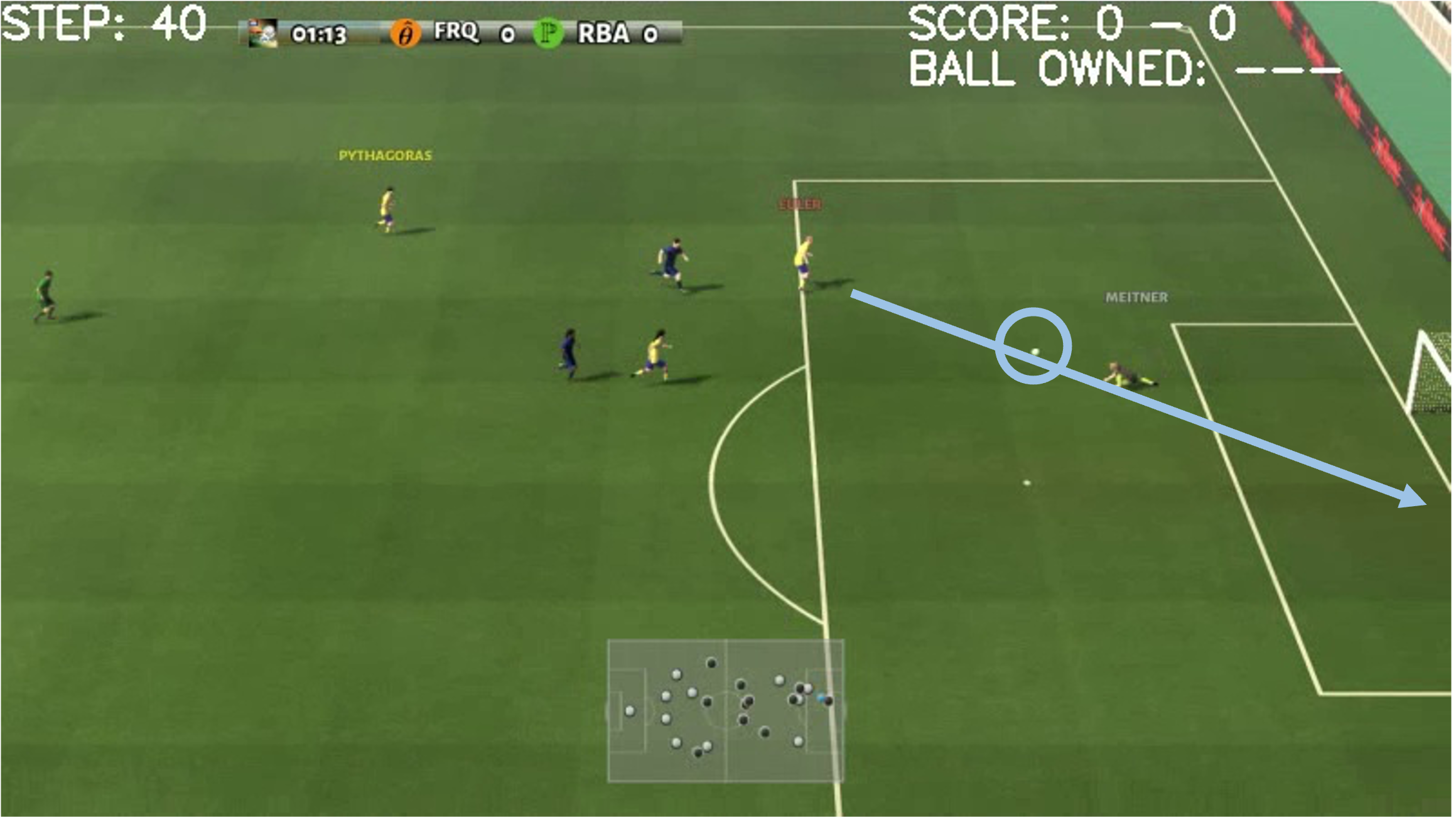}}
% \subfigure[3\_vs\_1\_with\_keeper]{
%         \includegraphics[width=0.23\textwidth]{figs/results/vis_1.png}
%         \label{fig:vis1}
%     }
% \subfigure[corner]{
%         \includegraphics[width=0.23\textwidth]{figs/results/vis_2.png}
%         \label{fig:vis2}
%     }
% \subfigure[counterattack\_hard]{
%         \includegraphics[width=0.23\textwidth]{figs/results/vis_3.png}
%         \label{fig:vis3}
%     }
% \subfigure[3m]{
%         \includegraphics[width=0.23\textwidth]{figs/results/vis_4.png}
%         \label{fig:vis4}
%     }

\caption{Visualization in the counterattack\_hard scenario. We visualize some key frames of trained policies, with our team in yellow. The yellow dashed arrows denote the player movement while the blue arrows denote the ball movement. On the left bottom corners, we visualize the intrinsic scaffolds of the agent holding the ball, and red bars denote actions encouraged by intrinsic scaffolds. \looseness=-1}
\label{fig:vis}
% \vskip -0.1in
\end{figure*}

\subsection{Ablation Studies}\label{sec: ablations}
We conduct three sets of ablation studies regarding different aspects of our proposed ICES with one representative task from each benchmark task.

The first set of ablation studies focuses on individual contributions. We explore two distinct modifications to our original approach with the results presented in \cref{fig:ablation_ind}:

% - ICES with global contribution (ICES w/ global-con): Instead of using the individual contribution as scaffolds for the corresponding agent following \cref{eq: individual_scaffolds}, we use the collective contribution of all agents as scaffolds. Here, the contribution of all agents is estimated by $r_{t, \text{int}} = I(z_{t+1}; \boldsymbol{u}_t | s_t).$\\
% - ICES with combined intrinsic rewards and extrinsic rewards (ICES w/ int+ext): In this variant, individual scaffolds are summed up and used as global intrinsic rewards, similar to previous methods~\citep{CDS}.

- \textbf{ICES w/ global-con:} Instead of using the individual contribution as scaffolds for the corresponding agent following \cref{eq: individual_scaffolds}, we use the collective contribution of all agents as a global scaffold. Here, the contribution of all agents is estimated by $r_{t, \text{int}} = I(z_{t+1}; \boldsymbol{u}_t | s_t).$\\
- \textbf{ICES w/ int+ext:} In this variant, individual scaffolds are directly summed up and used as global intrinsic rewards, similar to previous methods~\citep{CDS}.

% Results in \cref{fig:ablation_ind} indicate that replacing individual contributions with global contributions hinders effective exploration. This is supported by the increased number of training timesteps required to develop effective policies. We can see this is especially evident in the task \textit{3\_vs\_1\_with\_keeper}, where ICES achieves the final performance of over 60\% winning rate while ICES w/ global-con only archives 40\%. This is because the contribution is clearly mainly made by the agent with the ball, therefore equally assigning intrinsic scaffolds to all agents will results in misattribution. Further integrating the scaffolds into a global intrinsic reward exacerbates performance degradation. This could be attributed to the added complexity of needing to assign credit among agents of this new, non-stationary intrinsic reward, complicating the training process. Thus, this set of ablation studies underscores the effectiveness of directly assigning individual scaffolds to agents.
Results in \cref{fig:ablation_ind} indicate that replacing individual contributions with global contributions hinders effective exploration. Notably, in the \texttt{3\_vs\_1\_with\_keeper}
task, ICES achieves the final performance of over 60\% winning rate while ICES w/ global-con only archives 40\%. This highlights the misallocation of exploration incentives when contributions are not individually attributed, particularly in scenarios where specific agents play pivotal roles (e.g. the player with the ball in GRF). Moreover, further integrating the scaffolds into a global intrinsic reward exacerbates performance degradation. This could be attributed to the added complexity of needing to assign credit among agents of this new, non-stationary intrinsic reward, complicating the training process. Thus, this set of ablation studies underscores the effectiveness of directly assigning individual scaffolds to agents.

The second set of ablation studies investigates the effect of decoupling exploration and exploitation policies. We conducted experiments with two variants with the results shown in \cref{fig:ablation_entangled}:

- \textbf{ICES w/o s:} This variant, diverging from the approach specified in \cref{eq: exploration_policy}, excludes the global observation $s$ from the exploration policy's inputs. \\
- \textbf{ICES w/ int+ext:} In this variant, individual scaffolds are directly summed up and used as global intrinsic rewards, similar to previous methods~\citep{CDS}.
% - ICES without global states as inputs for exploration policy (ICES w/o s): Diverging from the approach specified in \cref{eq: exploration_policy}, this variant omits the global state $s$ from the exploration policy's inputs. \\
% - ICES with combined intrinsic reward and extrinsic reward (ICES w/ int+ext)

\cref{fig:ablation_entangled} shows the detrimental impact on exploration effectiveness when excluding global observation information from the exploration policies (while maintaining separate networks for exploration and exploitation). This highlights the significance of utilizing privileged information in exploration policies. Particularly in scenarios with pronounced partial observability, such as those encountered in the SMAC tasks, the lack of global information heavily deteriorates the exploration, with ICES achieving a final winning rate of 60\% while ICES w/o s only achieving 20\%. Furthermore, the performance further degrades when exploration and exploitation policies are merged into a single network. This set of ablation studies emphasizes the critical role of decoupling exploration and exploitation policies.

The last set of ablations are for other design choices in ICES and the results are given in \cref{fig:ablation_others}:

- \textbf{ICES w/o max-ent:} This variation eliminates the entropy regularization by by setting $\beta = 0$ in \cref{eq: obj_int}. \\
- \textbf{ICES w/o CVAEs:} Contrary to leveraging KL-divergence in the latent space as stated in \cref{eq: individual_scaffolds}, this variant directly calculates the intrinsic contribution as the Euclidean distance in the original state space. \\
- \textbf{ICES w/ 2 CVAEs:} Instead of employing two encoders and a shared decoder, this variant trains two independent CVAEs for Bayesian surprise estimation.

\cref{fig:ablation_others} indicates that each of these modifications leads to a decline in performance in terms of final performance or convergence speed. 

% - \textbf{ICES w/o max-ent:} We set $\beta = 0$ in \cref{eq: obj_int} for a variant of ICES without max entropy regularization. \\
% - ICES without CVAEs (ICES w/o CVAEs): We calculate the intrinsic contribution using the Euclidean distance in the original state space instead of KL-divergence in the latent space as stated in \cref{eq: individual_scaffolds}. \\
% - ICES with two CVAEs (ICES w/ 2 CVAEs): We train two independent CVAEs instead of two encoders and a shared decoder for Bayesian surprise estimation.

\subsection{Hyperparameter Analysis}
We further investigate the effect of different hyperparameters on the performance of ICES, as shown in \cref{fig:hyper_ana}. The hyperparameter $\alpha$ controls the tradeoff between the exploration policy and exploitation policy, while $\beta$ determines the balance between random exploration and directed exploration. Overall, within a reasonable range, ICES performs competitively across different hyperparameter settings, showcasing its robustness. However, achieving optimal performance requires proper tuning based on the specific task at hand.

\subsection{Visualization}

We visualize the final trained policies alongside some intrinsic scaffolds of the \texttt{counterattack\_hard} scenario in \cref{fig:vis}. We observe that on the first few timesteps, player 8, who possesses the ball, moves towards the right, aiming to approach its teammate palyer 7. At the same time, one of the highest rewarded actions identified by the intrinsic scaffolds is \textit{short\_pass}, which is beneficial because player 7 is closer to the goal. Consequently, guided by this intrinsic scaffold, player 8 executes a pass to player 7. Subsequently, player 7 receives the pass and makes a shot, resulting in a goal. Notably, right before the goal, player 7 is encouraged to \textit{shoot} or \textit{sprint}, both of which are good action candidates. This visualization result showcases how intrinsic scaffolds serve as a guiding mechanism, directing agents towards actions that are both exploratory and strategically sound.

\section{Conclusions}
In this work, we investigate MARL with sparse rewards. To facilitate cooperative exploration among agents without tampering the training objective, we propose ICES. Its key idea is to use estimations of individual contributions to encourage agents to choose actions that have more significant impact on the latent state transition during training time. ICES offers two main benefits: Firstly, with the individual contribution estimated by Bayesian surprise, ICES directly assigns the exploration credits to individual agents. This approach bypasses the need for credit assignment of global intrinsic rewards and alleviates the noisy TV problem brought by stochastic environment transitions. Secondly, with the distinct algorithms and objectives to optimize exploration and exploitation policies, ICES retains the original MARL goal of maximizing extrinsic rewards while enjoying the benefit of cooperative exploration. 

This paper has two main limitations. First, the proposed ICES requires additional policy networks, which introduce extra training complexity compared with QMIX or QPLEX. Reducing such complexity might be worth exploring when scaling ICES to cases with more agents.
Second, this paper only considers one-step latent state transitions, which may be insufficient as exploration guidance in more complicated scenarios. For future work, we aim to incorporate time abstraction in ICES to further improve its applicability.

\section*{Acknowledgements}

We thank the anonymous reviewers for their valuable feedback and suggestions.

\section*{Impact Statement}

This paper presents work whose goal is to advance the field of Reinforcement Learning. There are many potential societal consequences of our work, none of which we feel must be specifically highlighted here.

\nocite{langley00}

\bibliography{bibliography}
\bibliographystyle{icml2024}

%%%%%%%%%%%%%%%%%%%%%%%%%%%%%%%%%%%%%%%%%%%%%%%%%%%%%%%%%%%%%%%%%%%%%%%%%%%%%%%
%%%%%%%%%%%%%%%%%%%%%%%%%%%%%%%%%%%%%%%%%%%%%%%%%%%%%%%%%%%%%%%%%%%%%%%%%%%%%%%
% APPENDIX
%%%%%%%%%%%%%%%%%%%%%%%%%%%%%%%%%%%%%%%%%%%%%%%%%%%%%%%%%%%%%%%%%%%%%%%%%%%%%%%
%%%%%%%%%%%%%%%%%%%%%%%%%%%%%%%%%%%%%%%%%%%%%%%%%%%%%%%%%%%%%%%%%%%%%%%%%%%%%%%
\newpage
\appendix
\onecolumn
\section{ICES Algorithm Details} 
\subsection{ICES Exploration Policy Gradient} \label{sec: appendix_exploration_gradient}
Here we detail the deviation to \cref{eq: exploration_gradient}. From \cref{eq: obj_int}, we have: 
\begin{align}
    \mathcal{J}_i(\xi) = \mathbb{E}_{\nu_i}[r^i_\text{int}] + \beta \mathcal{H}(\cdot |\tau^i, s).
\end{align}
Then, we introduce $V_\eta (\tau^i, s)$ as a baseline. Since $V_\eta (\tau^i, s)$ is independent to $\nu_i$, we can rewrite the objective as:
\begin{align}
    \mathcal{J}_i(\xi) &= \mathbb{E}_{\nu_i(\xi)}\left[r^i_\text{int} - V_\eta (\tau^i, s)\right] + \beta \mathcal{H}(\cdot |\tau^i, s) \\
    &= \mathbb{E}_{s,\tau^i \sim d, \nu_i(\xi)}\left[R^i_\text{int}(s, a) - V_\eta (\tau^i, s)\right] + \beta \mathcal{H}(\cdot |\tau^i, s),
\end{align}
where $d$ is the distribution of $\tau^i$ and $s$, and $R^i_\text{int}(\cdot, \cdot)$ is the intrinsic reward function.

Taking the gradient of $\mathcal{J}_i(\xi)$ with respect to $\xi$, we have:
\begin{align}
    \nabla_\xi \mathcal{J}_i(\xi) = \nabla_\xi \mathbb{E}_{s,\tau^i \sim d, \nu_i(\xi)}\left[R^i_\text{int}(s, a) - V_\eta (\tau^i, s)\right] + \beta \nabla_\xi \mathcal{H}(\cdot |\tau^i, s).
\end{align}
Define $r_{sa} = \mathbb{E} \left[ R^i_\text{int}(s, a) - V_\eta (\tau^i, s) | s=s, a=a \right],$ we can rewrite the gradient as: 
\begin{align}
    \nabla_\xi \mathcal{J}_i(\xi) &= \nabla_\xi \sum_{\tau^i, s} d(\tau^i, s) \sum_a \nu_{i, \xi} (a|s) r_{sa} + \beta \nabla_\xi \mathcal{H}(\cdot |\tau^i, s)\\
    &= \sum_{\tau^i, s} d(\tau^i, s) \sum_a r_{sa} \nabla_\xi \nu_{i, \xi} (a|\tau^i, s) + \beta \nabla_\xi \mathcal{H}(\cdot |\tau^i, s)\\
    &= \sum_{\tau^i, s} d(\tau^i, s) \sum_a r_{sa} \cdot \nu_{i, \xi}(a|\tau^i, s) \frac{\nabla_\xi  \nu_{i, \xi}(a|\tau^i, s)}{\nu_{i, \xi}(a|\tau^i, s)} + \beta \nabla_\xi \mathcal{H}(\cdot |\tau^i, s)\\
    &= \sum_{\tau^i, s} d(\tau^i, s) \sum_a \nu_{i, \xi}(a|\tau^i, s) \cdot r_{sa} \nabla_\xi \ln{\nu_{i, \xi}(a|\tau^i, s)} + \beta \nabla_\xi \mathcal{H}(\cdot |\tau^i, s)\\
    &= \mathbb{E}_{s,\tau^i \sim d, \nu_i(\xi)} \left[ \left( R^i_\text{int}(s, a) - V_\eta (\tau^i, s) \right) \cdot \ln{\nu_{i, \xi}(a|\tau^i, s)} \right]+ \beta \nabla_\xi \mathcal{H}(\cdot |\tau^i, s)\\
    &= \mathbb{E}_{s,\tau^i \sim d, \nu_i(\xi)} \left[ \left( R^i_\text{int}(s, a) - V_\eta (\tau^i, s) - \beta \right) \cdot \ln{\nu_{i, \xi}(a|\tau^i, s)} \right].
\end{align}

\subsection{ICES Training Details} \label{sec: appendix_training_details}
We detail the training procedures for the policy networks and the scaffolds network mentioned in \cref{sec: algo_pseudo_codes} with \cref{algo: TrainPolicies} and \cref{algo: TrainScaffolds}, respectively.
\begin{algorithm}[h]
\caption{TrainPolicies: Training Procedure of ICES Policies (\cref{sec: MARL_training})}
\begin{algorithmic}[1]
% \STATE $\text{batch} = \{s_k, \boldsymbol{o}_k, \boldsymbol{u}_k, r_k, s_{k+1}, \boldsymbol{o}_{k+1}\}_{k=1}^{\text{batch\_size}} \sim \mathcal{D}$ 
\STATE {\bfseries Input:} Scaffolds parameters $\psi, \phi$, exploration network parameters $\xi, \eta$, exploitation networks parameters $\zeta$, replay buffer $\mathcal{D}$
\STATE Sample $\text{batch}  \sim \mathcal{D}$ 
% \hfill \COMMENT{$\triangleright$ Training policies. }
\STATE $\zeta \leftarrow \zeta - \text{LearningRate} \cdot \nabla \mathcal{L}(\zeta)$ 
\hfill \COMMENT{$\triangleright$ \cref{eq: obj_ext}}
\FOR{$i = 1, 2, ..., n$}
    \STATE Calculate intrinsic scaffolds 
    % \COMMENT{$\triangleright$ \cref{eq: individual_scaffolds}}
        $r^i_{t, \text{int}} = D_{\text{KL}} \left[ p_\psi(z_{t+1}|s_t, \boldsymbol{u}_t) \parallel p_\phi(z_{t+1}|s_t, \boldsymbol{u}_t^{-i}) \right]$
    \hfill \COMMENT{$\triangleright$ \cref{eq: individual_scaffolds}}
\ENDFOR
\STATE $\xi \leftarrow \xi + \text{LearningRate} \cdot  \sum_{i}^n \nabla \mathcal{J}_i(\xi)$
\hfill \COMMENT{$\triangleright$ \cref{eq: obj_int}}
\STATE $\eta \leftarrow \eta - \text{LearningRate} \cdot \nabla \mathcal{L}(\eta)$
\hfill \COMMENT{$\triangleright$ \cref{eq: obj_int_v}}

% \STATE Calculate intrinsic scaffolds $\{r^i_{t, \text{int}}\}_{i=1}^n$ online following \cref{eq: individual_scaffolds}
% \STATE Update parameters for exploration networks $\xi, \eta$ according to \cref{eq: obj_int} and \cref{eq: obj_int_v}, based on $\{r^i_{t, \text{int}}\}_{i=1}^n$
\STATE {\bfseries Output:} Updated parameters $\xi, \eta, \zeta$
\end{algorithmic}
\label{algo: TrainPolicies}
\end{algorithm}

\begin{algorithm}[h]
\caption{TrainScaffolds: Training Procedure of ICES Scaffolds (\cref{sec: scaffolds_construction})}
\begin{algorithmic}[1]
\STATE {\bfseries Input:} Scaffolds parameters $\psi, \phi, \theta$, replay buffer $\mathcal{D}$
% \STATE Sample $\text{batch} = \{s_k, \boldsymbol{o}_k, \boldsymbol{u}_k, r_k, s_{k+1}, \boldsymbol{o}_{k+1}\}_{k=1}^{\text{batch\_size}} \sim \mathcal{D}$
\STATE Sample $\text{batch}  \sim \mathcal{D}$
% \hfill \COMMENT{$\triangleright$ Training scaffolds. (\cref{sec: scaffolds_construction})}
% \STATE Update scaffolds parameters $\psi, \phi, \theta$ utilizing global information $\{s_k, \boldsymbol{u}_k, s_{k+1}\}_{k=1}^{\text{batch\_size}}$ in the sampled batch, according to \cref{eq: obj_scaffolds}
\STATE $\psi \leftarrow \psi + \text{LearningRate} \cdot \nabla \mathcal{J}_\psi(\psi , \phi , \theta)$
% \hfill \COMMENT{$\triangleright$ \cref{eq: obj_scaffolds}}
\STATE $\phi \leftarrow \phi + \text{LearningRate} \cdot \nabla \mathcal{J}_\phi(\psi , \phi , \theta)$
% \hfill \COMMENT{$\triangleright$ \cref{eq: obj_scaffolds}}
\STATE $\theta \leftarrow \theta + \text{LearningRate} \cdot \nabla \mathcal{J}_\theta(\psi , \phi , \theta)$
\\\hfill \COMMENT{$\triangleright$ \cref{eq: obj_scaffolds}}
\STATE {\bfseries Output:} Updated parameters $\psi, \phi, \theta$
\end{algorithmic}
\label{algo: TrainScaffolds}
\end{algorithm}

\section{Experiment Details} \label{sec: appendix_implementation_details}

\subsection{Environmental Settings} \label{sec: appendix_env}

\textbf{Google Research Football (GRF): } We evaluate our proposed method ICES against baselines on three GRF~\citep{GRF} scenarios, namely \texttt {academy\_3\_vs\_1\_with\_keeper}, \texttt{academy\_corner} and \texttt{academy\_counterattack\_hard}. The sparse reward settings are used for ICES, baselines, and ablations, where the rewards are only observed when scoring or losing the game~\citep{CDS}. The details of the reward setting is given in \cref{tab: grf_rewards}. This reward structure calls for high levels of cooperation among agents and is further complicated by the stochastic nature of opponents’ policies. For GRF tasks (\cref{fig:3v1,fig:corner,fig:cah}), we plot the average scores (with $1$ for wining, $0$ for a tie and $-1$ for losing) of test episodes with respect to the training timesteps.

\begin{table}[th]
\caption{GRF rewards.}
\label{tab: grf_rewards}
\begin{center}
\begin{tabular}{lc}
\toprule
Event                                          & Reward \\
\midrule
Our team scores                                & +100   \\
Opponent team scores                           & -1     \\
Our team or the ball returns to our half-court & -1    \\
\bottomrule
\end{tabular}
\end{center}
\end{table}

\textbf{StarCraft Multi-agent Challenge (SMAC): } We further assess our proposed method ICES on five SMAC~\citep{SMAC} scenarios with sparse reward settings following the previous works~\citep{MASER, ADER}. The rewards are only given upon the death of units (allies or enemies), and details are listed in \cref{tab: smac_rewards}.  We use four easy tasks and one hard task for benchmark, including \texttt{3m}, \texttt{8m}, \texttt{2s3z}, \texttt{2s\_vs\_1sc} and \texttt{5m\_vs\_6m}, as specified in \cref{tab: smac_scenarios}. For SMAC tasks (\cref{fig:3m,fig:8m,fig:2s3z,fig:2s_vs_1sc,fig:5m_vs_6m}), we plot the average win rate of test episodes over training timesteps.

\begin{table}[th]
\caption{SMAC challenges.}
\label{tab: smac_scenarios}
\begin{center}
% \begin{sc}
\begin{tabular}{lccc}
\toprule
Task        & Ally Units              & Enemy Units             & Type                      \\ 
\midrule
3m          & 3 Marines               & 3 Marines               & homogeneous, symmetric    \\ 
8m          & 8 Marines               & 8 Marines               & homogeneous, symmetric    \\ 
2s3z        & 2 Stalkers \& 3 Zealots & 2 Stalkers \& 3 Zealots & heterogeneous, symmetric  \\ 
2s\_vs\_1sc & 2 Stalkers              & 1 Spine Crawler         & homogeneous, asymmetric   \\ 
5m\_vs\_6m  & 5 Marines               & 6 Marines               & heterogeneous, asymmetric \\ 
\bottomrule
\end{tabular}
% \end{sc}
\end{center}
\end{table}

\begin{table}[th]
\caption{SMAC rewards.}
\label{tab: smac_rewards}
\begin{center}
\begin{tabular}{lc}
\toprule
Event           & Reward \\
\midrule
All enemies die & +200   \\
One enemy dies  & +10    \\
One ally dies   & -5   \\
\bottomrule
\end{tabular}
\end{center}
\end{table}

\subsection{ICES Implementation Details} \label{sec: appendix_env}
For both ICES and baselines, parameter sharing among agents is adopted to improve the sample efficiency as well as lower the model complexity. For ICES specifically, we remove the $\epsilon$-greedy exploration strategy after the epsilon annealing time, since the exploration policy is already random in its nature.

For GRF, we implement ICES based on the code framework PyMARL~\citep{SMAC}. For SMAC, we implement ICES based on the code framework PyMARL 2~\citep{pymarl2}. In this environment, before agents conduct any meaningful exploration, they tend to prolong the episode by escaping instead of attacking the enemies. Previous methods avoid this behavior by normalizing the intrinsic rewards to be less than or equal to zero~\citep{MASER, fox}, while we simply add a $-0.02$ step penalty as intrinsic rewards. For both GRF and SMAC experiments, default hyperparameters from the code frameworks are used. For ICES-specific hyperparameters, we list them in \cref{tab: ICES_hyper}. In particular, $\alpha$ anneals gradually throughout the training process. 

\begin{table}[th]
\caption{ICES hyperparameters.}
\label{tab: ICES_hyper}
\begin{center}
\begin{tabular}{lclc}
\toprule
Hyperparameter           & Benchmark  & Scenario           &Value \\
\midrule
Action embedding dimension & - & - & 4   \\
Scaffolds learning rate & - & - & 0.0001   \\
Scaffolds gradient clipping & - & - & 0.1   \\
\midrule
\multirow{2}{*}{Exploration agent learning rate} & GRF & - & 0.001   \\
  & SMAC & - & 0.01   \\
\midrule
\multirow{5}{*}{$\alpha$}  & \multirow{3}{*}{GRF} & {academy\_3\_vs\_1\_with\_keeper} & 0.2 - 0.05\\
  &  & {academy\_corner} & 0.2 - 0.05 \\
  &  & {academy\_counterattack\_hard} & 0.1 - 0.05 \\
& \multirow{2}{*}{SMAC} & {5m\_vs\_6m} & 0.1 - 0.05\\
  &  & others & 0.1 - 0.1 \\
\midrule
\multirow{5}{*}{$\beta$}  & \multirow{3}{*}{GRF} & {academy\_3\_vs\_1\_with\_keeper} & 0.02\\
  &  & {academy\_corner} & 0.05 \\
  &  & {academy\_counterattack\_hard} & 0.05 \\
& \multirow{2}{*}{SMAC} & {5m\_vs\_6m} & 0.5\\
  &  & others & 0.1 \\
\bottomrule
\end{tabular}
\end{center}
\end{table}

\subsection{Infrastructure}
Experiments are carried out on NVIDIA GeForce RTX 3080 GPUs.
%%%%%%%%%%%%%%%%%%%%%%%%%%%%%%%%%%%%%%%%%%%%%%%%%%%%%%%%%%%%%%%%%%%%%%%%%%%%%%%
%%%%%%%%%%%%%%%%%%%%%%%%%%%%%%%%%%%%%%%%%%%%%%%%%%%%%%%%%%%%%%%%%%%%%%%%%%%%%%%

\section{Additional Experimental Results} \label{sec: appendix_more_exp}
\subsection{ICES on KAZ}
\begin{figure}[ht]
% \vskip 0.2in
\begin{center}
\centerline{\includegraphics[scale=0.4]{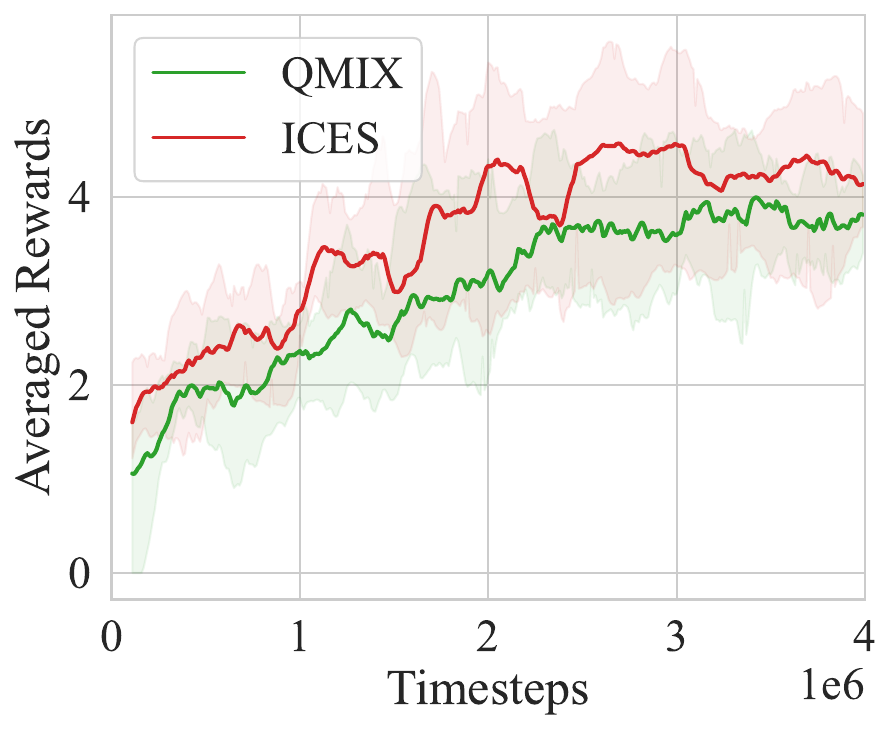}}
\caption{Performance comparison on KAZ benchmark.}
\label{fig:kaz}
\end{center}
% \vskip -0.2in
\end{figure}

Since ICES do not require particular parsing of states, it can be easily generalized to pixel-based MARL tasks. Therefore, we further test ICES on a pixel-based MARL benchmark task \texttt{knights\_archers\_zombies} (KAZ) from the pettingzoo environments~\citep{pettingzoo} with the results shown in \cref{fig:kaz}. We can see that in pixel-based MARL tasks, ICES is also able to improve the performance by promoting cooperative exploration.

\textbf{Knights Archers Zombies (KAZ) Environment: } In this game, we control 4 agents (2 knights and 2 archers) with the goal to kill all zombies that appear on the screen. Each agent can move and attack to kill zombies. When a knight attacks, it swings a mace in an arc in front of its current heading direction. When an archer attacks, it fires an arrow in a straight line in the direction of the archer’s heading. We reward the agents with $+1$ when a zombie dies. 

\textbf{Inputs Preprocessing: } We use the pixel-based local and global obaservation in this environment for ICES and QMIX. In particular, the global observation is represented by a $720 \times 1280 \times 3$ pixel colored image, while the observation of each agent is represented as a $512 \times 512 \times 3$ pixel colored image around the agent. As the input space is too large for RL algorithms to learn efficiently, we adopt some necessary preprocessing to local and global observation to reduce the dimension of input space.

The preprocessing pipeline for local observation is as follows: Color Reduction (Only take the first color channel and discard the rest) $\rightarrow$ Resize to $64 \times 64$. The preprocessing pipeline for the global observation is as follows: Color Reduction $\rightarrow$ Crop the central $720 \times 1100$ pixels $\rightarrow$ Resize to $128 \times 128$. 

\textbf{Network Architecture: } We use the code framework PyMARL 2~\citep{pymarl2} for this experiment. Since the inputs are images rather than vectors, we add feature encoding blocks in the agent network and the mixing hypernetwork to process the input images. Details are provided in \cref{tab: kaz_agent_encoder} and \cref{tab: kaz_mixer_encoder}, respectively. RNN layers are not used for this experiment to avoid extensive GPU memory consumption.

\begin{table}[th]
\caption{Observation encoding blocks in agent network.}
\label{tab: kaz_agent_encoder}
\begin{center}
\begin{tabular}{ccc}
\toprule
Layer & Operator                & \# Channels \\
\midrule
1         & Conv $3 \times 3$ \& MaxPooling & 32          \\
2         & Conv $3 \times 3$ \& MaxPooling & 64          \\
3         & Conv $3 \times 3$ \& MaxPooling & 128         \\
4         & Flatten \& FC          & 128  \\
\bottomrule
\end{tabular}
\end{center}
\end{table}

\begin{table}[th]
\caption{Observation encoding blocks in mixing hypernetwork.}
\label{tab: kaz_mixer_encoder}
\begin{center}
\begin{tabular}{ccc}
\toprule
Layer & Operator                & \# Channels \\
\midrule
1         & Conv $5 \times 5$ \& MaxPooling & 32          \\
2         & Conv $5 \times 5$ \& MaxPooling & 64          \\
3         & Conv $5 \times 5$ \& MaxPooling & 128         \\
4         & Flatten \& FC          & 128  \\
\bottomrule
\end{tabular}
\end{center}
\end{table}

For ICES, we use the same QMIX network for value functions and the CVAEs with network structure specified in \cref{tab: kaz_ices_encoder} and \cref{tab: kaz_ices_decoder}. We design the network architecture for CVAE here based on \url{https://github.com/AntixK/PyTorch-VAE}.

\begin{table}[th]
\caption{CVAE encoder in ICES.}
\label{tab: kaz_ices_encoder}
\begin{center}
\begin{tabular}{ccc}
\toprule
Layer & Operator                & \# Channels \\
\midrule
1         & Conv $5 \times 5$ & 8          \\
2         & Conv $5 \times 5$ & 16          \\
3         & Conv $5 \times 5$ & 32         \\
4         & Conv $3 \times 3$ & 64         \\
5         & Conv $3 \times 3$ & 128         \\
6         & Flatten \& FC          & 128  \\
7         & FC $\times 2$          & 64  \\
\bottomrule
\end{tabular}
\end{center}
\end{table}

\begin{table}[th]
\caption{CVAE decoder in ICES.}
\label{tab: kaz_ices_decoder}
\begin{center}
\begin{tabular}{ccc}
\toprule
Layer & Operator                & \# Channels \\
\midrule
1         & FC $\times 2$          & 64  \\
2         & TransposedConv $5 \times 5$ & 64          \\
3         & TransposedConv $5 \times 5$ & 32          \\
4         & TransposedConv $5 \times 5$ & 16         \\
5         & TransposedConv $5 \times 5$ & 8         \\
6         & TransposedConv $7 \times 7$ & 2         \\
\bottomrule
\end{tabular}
\end{center}
\end{table}

\textbf{Hyperparameters: } We use the default hyperparameters in PyMARL 2 except for $\text{batch\_size} = 4$ for both QMIX and ICES. For ICES specific hyperparameters, we list them in \cref{tab: kaz_ICES_hyper}.

\begin{table}[th]
\caption{ICES hyperparameters used in KAZ experiment.}
\label{tab: kaz_ICES_hyper}
\begin{center}
\begin{tabular}{lc}
\toprule
Hyperparameter           & Value \\
\midrule
Action embedding dimension & 4   \\
Scaffolds learning rate & 0.0001   \\
Scaffolds gradient clipping & 0.1   \\
Exploration agent learning rate & 0.001   \\
$\alpha$ & 0.1 - 0.05   \\
$\beta$ & 0.1   \\
\bottomrule
\end{tabular}
\end{center}
\end{table}
%%%%%%%%%%%%%%%%%%%%%%%%%%%%%%%%%%%%%%%%%%%%%%%%%%%%%%%%%%%%%%%%%%%%%%%%%%%%%%%
%%%%%%%%%%%%%%%%%%%%%%%%%%%%%%%%%%%%%%%%%%%%%%%%%%%%%%%%%%%%%%%%%%%%%%%%%%%%%%%

\end{document}